\documentclass[]{style/iromlab}

\usepackage{amsmath, amsfonts, amssymb, amsthm}
\usepackage{mathtools}
\usepackage{bm}
\usepackage{bbm}
\usepackage{units}

\usepackage{booktabs}
\usepackage{multirow}
\usepackage{tabularx}
\usepackage{longtable}
\usepackage{threeparttable}
\usepackage{adjustbox}

\usepackage{graphicx}
\usepackage{wrapfig}
\usepackage{float}
\usepackage{placeins}

\usepackage{enumerate}
\usepackage[inline]{enumitem}

\usepackage[ruled,algo2e]{algorithm2e}
\usepackage{algpseudocode}

\usepackage{xcolor}
\usepackage{tcolorbox}

\usepackage[numbers,sort&compress]{natbib}
\usepackage{cleveref}
\renewcommand{\backref}[1]{}
\renewcommand{\backrefalt}[4]{}

\usepackage{multicol}
\usepackage{caption}
\usepackage{balance}
\usepackage{microtype}

\usepackage{pifont}
\usepackage{etoolbox}
\usepackage{xspace}

\usepackage[title]{appendix}

\definecolor{claude_color}{HTML}{F89E62}
\definecolor{deepseek_color}{HTML}{78B6E8}
\definecolor{o3_mini_color}{HTML}{6CD5A1}
\definecolor{red_color}{HTML}{E13B55}
\definecolor{prompt_color}{HTML}{71502B}

\newcommand{\cmark}{\textcolor{green!60!black}{\ding{51}}}
\newcommand{\xmark}{\textcolor{red!70!black}{\ding{55}}}

\tcbset{
  promptbox/.style={
    colback=gray!6,
    colframe=gray!25,
    boxrule=0.4pt,
    arc=2pt,
    left=6pt,
    right=6pt,
    top=4pt,
    bottom=4pt
  }
}

\makeatletter
\newcommand{\longdash}[1][2em]{%
  \makebox[#1]{$\m@th\smash-\mkern-7mu\cleaders\hbox{$\mkern-2mu\smash-\mkern-2mu$}\hfill\mkern-7mu\smash-$}}
\makeatother

\author[1]{Lihan Zha}
\author[1*]{Asher J. Hancock}
\author[1*]{Mingtong Zhang}
\author[1]{Tenny Yin}
\author[1]{Yixuan Huang}
\author[1]{Dhruv Shah}
\author[2,\ensuremath{\dagger}]{Allen Z. Ren}
\author[1,\ensuremath{\dagger}]{Anirudha Majumdar}

\affiliation[1]{Princeton University}
\affiliation[2]{Physical Intelligence}
\contribution[*]{Equal contribution.}
\contribution[\ensuremath{\dagger}]{Equal advising.}

\begin{document}

\title{
LAP: Language-Action Pre-Training Enables \\ Zero-shot Cross-Embodiment Transfer
}

\abstract{
A long-standing goal in robotics is a generalist policy that can be deployed zero-shot on new robot embodiments without per-embodiment adaptation. Despite large-scale multi-embodiment pre-training, existing Vision--Language--Action models (VLAs) remain tightly coupled to their training embodiments and typically require costly fine-tuning. We introduce \emph{Language-Action Pre-training} (LAP), a simple recipe that represents low-level robot actions directly in natural language, aligning action supervision with the pre-trained vision--language model’s input--output distribution. LAP requires no learned tokenizer, no costly annotation, and no embodiment-specific architectural design. Based on LAP, we present \textsc{LAP-3B}, which to the best of our knowledge is the first VLA to achieve substantial zero-shot transfer to previously unseen robot embodiments without any embodiment-specific fine-tuning. Across multiple novel robots and manipulation tasks, \textsc{LAP-3B} attains over 50\% average zero-shot success, delivering roughly a $2\times$ improvement over the strongest prior VLAs. We further show that LAP enables efficient adaptation and favorable scaling, while unifying action prediction and VQA in a shared language-action format that yields additional gains through co-training.
}

\keywords{
Vision-Language-Action Models, Cross-Embodiment Learning
}

\website{
https://lap-vla.github.io/
}
{
https://lap-vla.github.io
}

\code{
https://github.com/lihzha/lap
}
{
https://github.com/lihzha/lap
}

\maketitle

\section{Introduction}
\label{sec:intro}

\begin{figure*}[t]
    \centering
    \includegraphics[width=\linewidth]{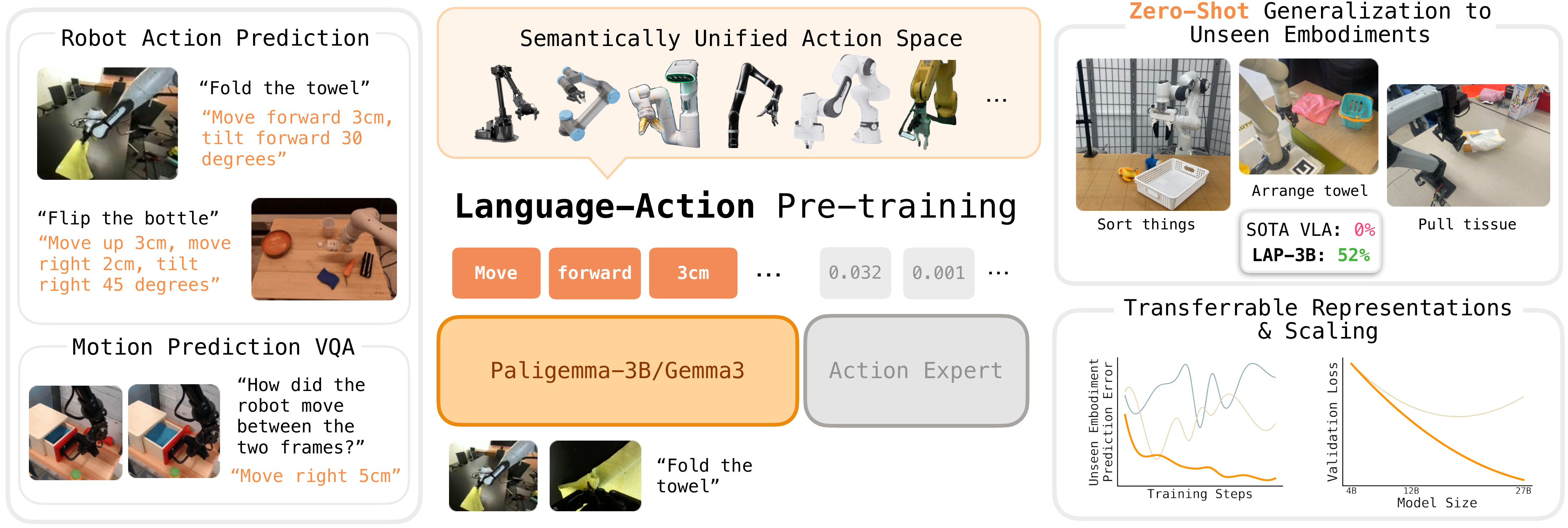}
    \caption{We introduce Language-Action Pre-training (LAP), a general VLA pre-training recipe that represents low-level actions directly in natural language to supervise a vision–language backbone, and instantiate it as \textsc{LAP-3B}, the first VLA to demonstrate strong \emph{zero-shot transfer} to novel embodiments. Compared to state-of-the-art VLAs, LAP-3B learns more generalizable embodiment representations and exhibits favorable scaling behavior.
        }
    \label{fig:teaser}
    
\end{figure*}

The ultimate promise of robot foundation policies is the ability to control \emph{any} robot embodiment to perform human-level intelligent tasks. Vision–Language–Action models (VLAs) have made remarkable progress toward this goal, leveraging pre-trained Vision–Language models (VLMs) to obtain strong visual–semantic priors for broad scene and even task generalization \cite{intelligence2025pi05visionlanguageactionmodelopenworld, lee2025molmoactactionreasoningmodels, geminiroboticsteam2025geminiroboticsbringingai, nvidia2025gr00tn1openfoundation, zheng2025xvlasoftpromptedtransformerscalable, goyal2025vla0buildingstateoftheartvlas, kim2024openvlaopensourcevisionlanguageactionmodel, kim2025finetuningvisionlanguageactionmodelsoptimizing, liu2025rdt1bdiffusionfoundationmodel}. However, despite large-scale multi-embodiment training \cite{embodimentcollaboration2025openxembodimentroboticlearning, khazatsky2025droidlargescaleinthewildrobot, agibotworldcontributors2025agibotworldcolosseolargescale, galaxea2025, wu2025robomindbenchmarkmultiembodimentintelligence}, state-of-the-art VLAs still rarely function \emph{zero-shot on new robots} with even minor differences to training embodiments (e.g., an altered gripper or wrist camera location), as shown in Fig.~\ref{fig:teaser}. In practice, deploying a VLA on a new robot often requires extensive per-embodiment adaptation, which is costly and undermines the goal of a single broadly deployable foundation model.

At its core, zero-shot cross-embodiment transfer is difficult because the embodiment space is vast and fundamentally expensive to cover with data. While task and environment diversity can scale via in-the-wild robot data \cite{chi2024universalmanipulationinterfaceinthewild, khazatsky2025droidlargescaleinthewildrobot, tao2025dexwilddexteroushumaninteractions} and internet-scale egocentric video \cite{grauman2022ego4dworld3000hours, liu2024tacobenchmarkinggeneralizablebimanual, banerjee2025hot3dhandobjecttracking, wang2023holoassistegocentrichumaninteraction, hoque2025egodexlearningdexterousmanipulation, damen2020epickitchensdatasetcollectionchallenges}, embodiment diversity remains constrained by hardware availability and data-collection logistics; for example, the Open X-Embodiment dataset aggregates data from only a few dozen robot types \cite{embodimentcollaboration2025openxembodimentroboticlearning}. Achieving zero-shot transfer therefore requires extracting maximal generalization from limited embodiment diversity, rather than relying on exhaustive coverage.

Our key observation is that zero-shot cross-embodiment transfer depends critically on how we adapt a pre-trained VLM for motor control. Concretely, the model must acquire precise control capabilities while retaining the semantic and visual representations that enable transfer across embodiments. When this balance is mismanaged, embodiment-specific action learning can override the very features needed for generalization. However, because VLMs are not pre-trained to interpret or generate motor-level, high-frequency control signals, standard VLA training can induce distributional mismatch and degrade pre-trained knowledge \cite{hancock2025actionslanguagefinetuningvlms, zhou-etal-2025-chatvla, driess2025knowledgeinsulatingvisionlanguageactionmodels}.

Motivated by this observation, we propose \emph{Language-Action Pre-training (LAP)}, a simple pre-training recipe for VLAs that adapts a pre-trained VLM backbone for motor control by training it to predict \emph{actions described in natural language}, herein referred to as \emph{language-actions}. Language-actions express the net effect of a raw end-effector action (e.g., ``move left 5 cm'', ``tilt forward 45 degrees'') in the same modality the VLM is trained to model, as shown in Fig.\ref{fig:teaser}. They can be parsed from raw actions under a \emph{fixed} coordinate convention and template without additional data annotation. 

This representation choice directly targets the distributional mismatch that arises in standard VLA training recipes, which fine-tune VLMs to predict \emph{raw actions}—continuous joint or end-effector commands, or discretizations thereof—represented by arbitrary or abstract tokens~\cite{kim2024openvlaopensourcevisionlanguageactionmodel, pertsch2025fastefficientactiontokenization, lee2025molmoactactionreasoningmodels}. Such representations carry little semantic structure that could unify behaviors across embodiments \cite{zheng2025universalactionsenhancedembodied, ye2025latentactionpretrainingvideos, bu2025univlalearningacttaskcentric, chen2025embodimentequivariantvisionlanguageactionpolicy}. Notably, this limitation persists even when a unified end-effector action space is adopted: existing VLAs still exhibit little to no zero-shot generalization to novel embodiments~\cite{kim2024openvlaopensourcevisionlanguageactionmodel, zheng2025xvlasoftpromptedtransformerscalable} (see Section~\ref{subsec:zero_shot}). Together, these results suggest that cross-embodiment transfer cannot be achieved by action space unification alone, but instead requires a fundamentally different action representation. Language-actions provide such a representation by keeping the supervision signal close to the VLM’s pre-training distribution \cite{hancock2025actionslanguagefinetuningvlms}, promoting semantically grounded features that transfer across embodiments.

With this representation in place, we next consider how to instantiate LAP in a practical VLA architecture. In principle, LAP can be applied to any VLM-based VLA by treating action generation as a language modeling problem. However, such formulations typically rely on auto-regressive language-action token sampling at inference time, which is inefficient and ill-suited for robot control \cite{kim2024openvlaopensourcevisionlanguageactionmodel, pertsch2025fastefficientactiontokenization, driess2025knowledgeinsulatingvisionlanguageactionmodels, hancock2025actionslanguagefinetuningvlms}. We therefore adopt a hybrid design that combines a LAP-trained VLM backbone with a lightweight diffusion-based action expert for continuous control, following $\pi_{0.5}$~\cite{intelligence2025pi05visionlanguageactionmodelopenworld}. This Mixture-of-Transformers architecture \cite{liang2024mixture} preserves the representation benefits of language-action supervision while enabling efficient execution.

\noindent\textbf{Statement of contributions.} We introduce Language-Action Pre-training (LAP), a general pre-training recipe for Vision--Language--Action models that represents actions as natural language to enable cross-embodiment generalization. We instantiate LAP in \textsc{LAP-3B}, which pairs a VLM backbone with a lightweight action expert for efficient real-time control. Trained on existing open-sourced robot datasets \cite{embodimentcollaboration2025openxembodimentroboticlearning, khazatsky2025droidlargescaleinthewildrobot, lee2025molmoactactionreasoningmodels}, \textsc{LAP-3B} achieves substantial zero-shot generalization to previously unseen robot embodiments, delivering roughly a 2× relative improvement (approximately 30\% absolute) over prior pre-training recipes based on alternative action representations. \emph{To the best of our knowledge, this is the first VLA system to operate zero-shot on novel real robot embodiments.}\footnote{All code, checkpoints, and data configurations are open-sourced at \href{https://lap-vla.github.io}{https://lap-vla.github.io}.}
In addition, \textsc{LAP-3B} can be fine-tuned to new embodiments and more complex dexterous tasks using significantly fewer demonstrations and gradient steps, matching prior performance with up to $2.5\times$ less data. Finally, we analyze why LAP learns more transferable embodiment representations and exhibits more stable training dynamics, demonstrate its favorable scaling behavior with increasing model size, and show that the language-action interface naturally supports co-training with auxiliary VQA objectives (e.g., motion prediction), yielding further performance gains.

\section{Related Work}
\label{sec:related_work}

\noindent\textbf{Vision-Language-Action Models.} Recent work has developed generalist robot policies trained on large and diverse robotic datasets \cite{embodimentcollaboration2025openxembodimentroboticlearning, khazatsky2025droidlargescaleinthewildrobot, ebert2021bridgedataboostinggeneralization, pmlr-v229-walke23a, fang2023rh20tcomprehensiveroboticdataset, dasari2020robonetlargescalemultirobotlearning, wu2025robomindbenchmarkmultiembodimentintelligence, pmlr-v164-jang22a, galaxea2025, agibotworldcontributors2025agibotworldcolosseolargescale}. Among these, Vision--Language--Action models (VLAs) have emerged as a particularly promising paradigm \cite{brohan2023rt1roboticstransformerrealworld, embodimentcollaboration2025openxembodimentroboticlearning, kim2024openvlaopensourcevisionlanguageactionmodel, octomodelteam2024octoopensourcegeneralistrobot, black2024pi0visionlanguageactionflowmodel, li2024cogactfoundationalvisionlanguageactionmodel, geminiroboticsteam2025geminiroboticsbringingai, liu2025rdt1bdiffusionfoundationmodel, brohan2023rt2visionlanguageactionmodelstransfer, nvidia2025gr00tn1openfoundation, hancock2025actionslanguagefinetuningvlms, pertsch2025fastefficientactiontokenization, wang2024scalingproprioceptivevisuallearningheterogeneous, zheng2025tracevlavisualtraceprompting, Zhao_2025_CVPR}. These models are typically fine-tuned from Vision--Language models (VLMs) pre-trained on internet-scale data. The pre-trained representations equip VLAs with the ability to follow natural language instructions and understand diverse visual cues. As a result, VLAs often not only perform well within their training distribution, but also show promise for generalization to out-of-distribution environments \cite{intelligence2025pi05visionlanguageactionmodelopenworld, black2024pi0visionlanguageactionflowmodel, kim2024openvlaopensourcevisionlanguageactionmodel, gao2024efficientdatacollectionrobotic, zha2025guidingdatacollectionfactored, hu2025datascalinglawsimitation}. Nevertheless, despite being trained on data spanning multiple robot embodiments, existing VLAs do not generalize zero-shot to novel embodiments without extensive fine-tuning (Section~\ref{subsec:zero_shot}). In this work, we introduce a new action representation that significantly improves VLA's embodiment generalization capability.

\vspace{5pt}
\noindent\textbf{Cross-Embodiment Policy Learning.} Large robot policies are typically trained on heterogeneous, cross-embodiment datasets, and a common strategy for handling varying state and action spaces is to manually define a unified representation. The most straightforward approach pads states and actions to the maximum dimensionality or uses shared action spaces like end-effector pose \cite{black2024pi0visionlanguageactionflowmodel, kim2024openvlaopensourcevisionlanguageactionmodel, zheng2025xvlasoftpromptedtransformerscalable}. Other works learn embodiment-specific state or action projectors, or employ separate action heads for different robot types \cite{wang2024scalingproprioceptivevisuallearningheterogeneous, doshi2024scalingcrossembodiedlearningpolicy, octomodelteam2024octoopensourcegeneralistrobot, yang2024pushinglimitscrossembodimentlearning, nvidia2025gr00tn1openfoundation}. A parallel line of work learns shared latent actions or skills across embodiments \cite{ye2025latentactionpretrainingvideos, bu2025univlalearningacttaskcentric, Zheng_2025_CVPR, Chen_2025_ICCV, xu2023xskillcrossembodimentskill, wang2024crossembodimentrobotmanipulationskill}. More recently, human hand motion has been used as a unified action space bridging human and robot data \cite{luo2026beingh05scalinghumancentricrobot,yang2025egovlalearningvisionlanguageactionmodels}, enabling task generalization to seen embodiments. While effective, these methods typically require embodiment-specific fine-tuning to adapt to new embodiments, whereas our method achieves \emph{zero-shot embodiment transfer} without any additional training.

Another line of work facilitates cross-embodiment learning by conditioning robot policies on embodiment features. Early approaches infer such features from kinematic structure \cite{huang2020policycontrolallshared, kurin2021bodycagerolemorphology, gupta2022metamorphlearninguniversalcontrollers, patel2024getzerographembodimenttransformer, bohlinger2025policyrunallendtoend}, and several works focus on human-to-robot transfer by exploiting structural similarity \cite{chen2024roviaugrobotviewpointaugmentation, lum2025crossinghumanrobotembodimentgap, lepert2025phantomtrainingrobotsrobots, ren2025motiontracksunifiedrepresentation}. For example, X-VLA \cite{zheng2025xvlasoftpromptedtransformerscalable} conditions policies on per-embodiment soft prompts, and DexVLA \cite{wen2025dexvlavisionlanguagemodelplugin} introduces an explicit embodiment-specific training stage. RDT2 \cite{liu2026rdt2exploringscalinglimit} similarly reports zero-shot embodiment transfer, but does so under the assumption that new robots share the same gripper design and wrist-mounted camera configuration as those used during data collection, effectively constraining embodiment variation to matched hardware interfaces rather than arbitrary morphologies. In parallel, a broader class of approaches infers embodiment implicitly from long interaction histories, as commonly explored in navigation and embodied control \cite{hirose2025omnivlaomnimodalvisionlanguageactionmodel, shah2023vintfoundationmodelvisual, yang2024pushinglimitscrossembodimentlearning, shah2023gnmgeneralnavigationmodel}. Taken together, these methods primarily model embodiment variation within the training distribution—through prompts, explicit embodiment conditioning, hardware alignment, or historical context—and typically require additional assumptions or adaptation when deployed on previously unseen embodiments.

By contrast, LAP is motivated by the complementary goal of true zero-shot embodiment transfer. Rather than introducing explicit embodiment-conditioning mechanisms or relying on matched hardware interfaces across robots (e.g., identical grippers and wrist-mounted cameras), LAP-3B demonstrates that cross-embodiment generalization can emerge directly from large-scale VLA pre-training with appropriately structured action representations. Without any embodiment-specific design or embodiment-aligned data collection assumptions, LAP-3B, to the best of our knowledge, is the first VLA to achieve substantial \emph{zero-shot generalization to previously unseen embodiments}, while also enabling efficient adaptation to more challenging tasks when additional data are available.

\vspace{5pt}
\noindent\textbf{Action Representations for VLAs.}  How actions are represented determines how effectively a pre-trained VLM can be adapted for motor control  \cite{lee2025molmoactactionreasoningmodels, pertsch2025fastefficientactiontokenization, liu2025fasterefficientautoregressivevision, goyal2025vla0buildingstateoftheartvlas, grover2025enhancinggeneralizationvisionlanguageactionmodels, niu2024llarvavisionactioninstructiontuning, ye2025latentactionpretrainingvideos, hancock2025actionslanguagefinetuningvlms}. Early approaches represent actions as code or language-level sub-tasks \cite{ahn2022icanisay, driess2023palmeembodiedmultimodallanguage, liang2023codepolicieslanguagemodel, zha2024distillingretrievinggeneralizableknowledge, belkhale2024rthactionhierarchiesusing}. These formulations are closely related in spirit to our use of linguistic supervision, but they typically operate at coarse temporal abstraction and rely on separate low-level controllers to execute each sub-task. In contrast, LAP uses \emph{fine-grained} language-actions parsed directly from continuous end-effector motion and trains the backbone to model these motor-level effects, yielding representations tailored for low-level control. More recent works instead encode actions as 2D keypoints or visual traces \cite{liu2024mokaopenworldroboticmanipulation, niu2024llarvavisionactioninstructiontuning, lee2025molmoactactionreasoningmodels, zheng2025tracevlavisualtraceprompting, dipalo2024keypointactiontokensenable, huang2024rekepspatiotemporalreasoningrelational}, which enhance grounding but are typically still used as intermediate representations rather than direct control.

\begin{wraptable}{r}{0.6\columnwidth}
\vspace{-6pt}
\centering
\scriptsize
\setlength{\tabcolsep}{3pt}
\renewcommand{\arraystretch}{0.9}
\begin{tabular}{lccc|c}
\hline
\vspace{2pt}
 & 
\citet{grover2025enhancinggeneralizationvisionlanguageactionmodels} & 
VLA-0 \cite{goyal2025vla0buildingstateoftheartvlas} & 
VLM2VLA \cite{hancock2025actionslanguagefinetuningvlms} & 
\textbf{LAP} \\
\hline
Large-Scale Training  & \cmark & \xmark & \xmark & \cmark \\
Inference Frequency  & $\sim$0.5Hz & 4Hz & 0.2Hz & \textbf{25Hz} \\
Cross-Embodiment     & \xmark & \xmark & \xmark & \cmark \\
\hline
\end{tabular}
\caption{Comparison of VLAs that use (variants of) language-actions along key design and performance axes. LAP uniquely enables cross-embodiment generalization through large-scale pre-training and language-action representation.}
\vspace{-8pt}
\label{tab:method_comparison_axes_transposed}
\end{wraptable}

Another line of work maps actions to unused or non-linguistic tokens in VLM vocabularies \cite{brohan2023rt2visionlanguageactionmodelstransfer, kim2024openvlaopensourcevisionlanguageactionmodel, pertsch2025fastefficientactiontokenization, liu2025fasterefficientautoregressivevision}, which can induce distributional misalignment and degrade pre-trained VLM representations \cite{hancock2025actionslanguagefinetuningvlms}. More recently, approaches have begun representing actions directly in natural language \cite{hancock2025actionslanguagefinetuningvlms} or as raw string tokens \cite{goyal2025vla0buildingstateoftheartvlas, grover2025enhancinggeneralizationvisionlanguageactionmodels}, yielding improved semantic grounding, language following, and task-level generalization. As summarized in Table~\ref{tab:method_comparison_axes_transposed}, LAP builds on this emerging paradigm with two key distinctions: it introduces LAP-3B as a \emph{practical} language-action VLA for real-time control, and demonstrates that language-action representations can scale to large robot pre-training corpora—thereby enabling zero-shot cross-embodiment transfer.

\begin{figure*}[t]
    \centering
    \includegraphics[width=\linewidth]{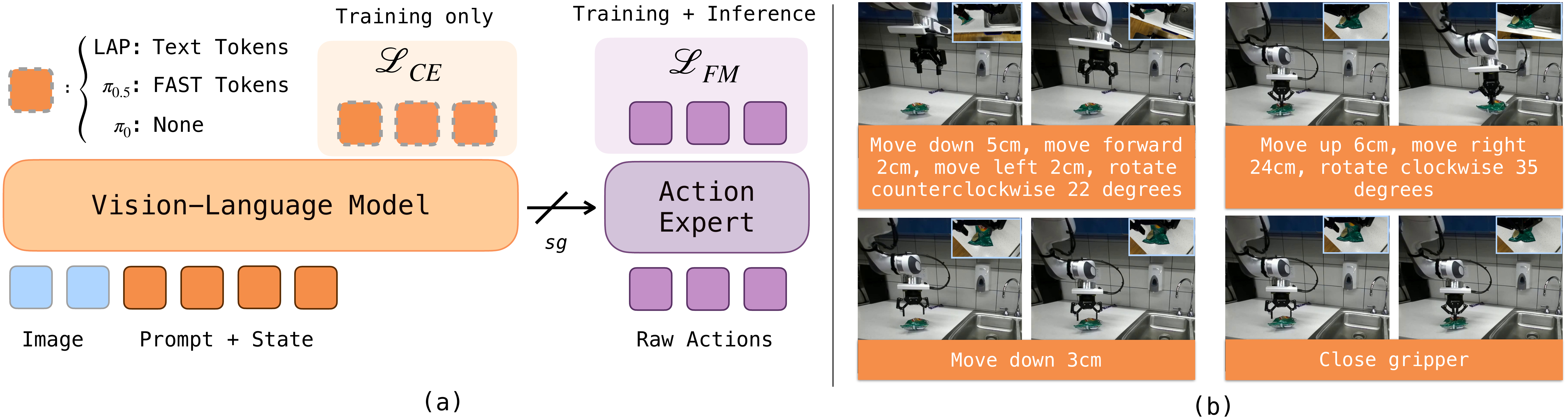}
    \caption{
        (a) A unified view comparing LAP-3B with prior VLAs in terms of action representation. A VLM backbone predicts discrete language-action tokens using a cross-entropy objective, while a lightweight action expert predicts continuous actions via flow matching. Gradients from the action expert are blocked from the VLM through knowledge insulation \cite{driess2025knowledgeinsulatingvisionlanguageactionmodels}, ensuring that the VLM is trained purely via language supervision. At test time, the action expert is rolled out for fast inference.
        (b) Visualizations of language-actions from the DROID dataset \cite{khazatsky2025droidlargescaleinthewildrobot}.
        }
    \label{fig:method}
\end{figure*}

\section{LAP: Language-Action Pre-Training for VLAs}
\label{sec:method}

We now introduce \emph{Language-Action Pre-training} (LAP), a general pre-training recipe for VLAs. We first formalize LAP's language-action learning objective (Section~\ref{subsec:problem_formulation}), then describe how to obtain language-actions at scale without manual annotation (Section~\ref{subsec:curation}). We then present \textsc{LAP-3B}, a vision-language-action model trained with LAP that demonstrates substantial cross-embodiment transfer (Section~\ref{subsec:architecture}), followed by specific implementation details (Section~\ref{sec:implementation_details}).

\subsection{Formulations}
\label{subsec:problem_formulation}

LAP reframes action learning in VLAs as a standard conditional language modeling problem. Given an observation $o_t = \{I_t^1, \dots, I_t^n, s_t\}$ consisting of multi-view RGB images and proprioceptive state, and a task instruction $l$, we deterministically derive a structured language-action sequence $\hat{a}_{t:t+H}^{\text{lang.}}$, as shown in Fig.\ref{fig:method}$(b)$, summarizing the corresponding continuous action chunk $a_{t:t+H}$.

The VLM backbone is trained to auto-regressively predict the language-action tokens conditioned on $(o_t, l)$ using the standard cross-entropy objective
\begin{equation}
\mathcal{L}_{\text{CE.}}
=
-\mathbb{E}_{(\hat{a}_{t:t+H}^{\text{lang.}}, o_t, l) \sim \mathcal{D}}
\sum_{i}
\log p_{\theta}
\big(
\hat{a}^{\text{lang.}}_{t,i}
\mid o_t, l, \hat{a}^{\text{lang.}}_{t,<i}
\big),
\label{eq:language-action}
\end{equation}
which is identical to standard sequence-to-sequence language modeling and corresponds to maximum likelihood estimation of language-actions, analogous to behavioral cloning for continuous actions.

\subsection{Language-Action Dataset Curation}
\label{subsec:curation}

A key requirement for LAP is the access to language-action supervision at scale. Rather than relying on manual annotation, we leverage the structured nature of robotic datasets to derive language-actions directly from end-effector trajectories.

Specifically, we represent language-actions as natural language descriptions of end-effector delta motions. Under a fixed coordinate convention and a small set of structured templates, continuous actions can be deterministically converted into language descriptions with arbitrary numerical resolution.

\noindent\textbf{Coordinate System.}
We adopt a convention commonly used in robotics datasets, where $+x$ is forward, $+y$ is left, and $+z$ is up; rotational components are expressed as Euler angles following the right-hand rule. Since most manipulation datasets already follow this convention, language-actions can be parsed directly without additional frame transformations. 

To encourage embodiment-agnostic reasoning and consistent use of multi-view visual observations, we additionally express language-actions in multiple reference frames. Specifically, during training we randomize the reference frame used to generate each language-action, expressing 50\% in the robot base frame and 50\% in the end-effector frame, and include the name of the reference frame in the model prompt. 

\noindent\textbf{Templates.}
Language-actions are generated using a small set of structured natural language templates that describe translational and rotational end-effector motions:
\begin{equation}
\hat{a}_t^{\text{lang.}}
=
\text{``}\langle\text{verb}\rangle\,
\langle\text{direction}\rangle\,
\langle\text{magnitude}\rangle\,
\langle\text{unit}\rangle\text{''},
\end{equation}
where the verb determines the action category: \emph{move} denotes translational end-effector displacement, \emph{tilt} denotes roll and pitch rotations, and \emph{rotate} denotes yaw rotation. For translation, directions correspond to $\{\pm x, \pm y, \pm z\}$ (e.g., forward/backward, left/right, up/down). For roll and pitch, we use ``left/right'' and ``back/forward'' following the right-hand rule convention. For yaw, ``clockwise'' and ``counterclockwise'' represent negative and positive rotations, respectively.

Rather than generating language-actions at every timestep, we represent $\hat{a}_{t:t+H}^{\text{lang.}}$ as a single description summarizing the net displacement of the action chunk $a_{t:t+H}$. Since actions are expressed as delta end-effector poses, this corresponds to the cumulative translation and rotation between timesteps $t$ and $t+H$. This temporal abstraction produces a low-frequency supervision signal that is easier for VLMs to model while abstracting away high-frequency control. We also discretize magnitudes to integer values and express translations in centimeters (e.g., \emph{``move forward 5 cm''}), yielding language-actions that are both structured and precise. Visual examples are shown in Fig.~\ref{fig:method}(b).

Compared to prior action tokenization approaches that discretize continuous actions into arbitrary or out-of-vocabulary symbols, language-actions remain well supported by the VLM’s natural language pre-training distribution, mitigating the distributional mismatch known to degrade representation quality \cite{hancock2025actionslanguagefinetuningvlms, driess2025knowledgeinsulatingvisionlanguageactionmodels}. In contrast to learned tokenizers such as FAST \cite{pertsch2025fastefficientactiontokenization}, which introduce a separate representation learning stage to compress actions into discrete non-linguistic tokens, language-actions are produced purely through deterministic parsing, eliminating the need for any learned action tokenizer. Moreover, unlike conventional action representations, language-actions naturally align with the input–output format of Visual--Question--Answering (VQA) tasks, enabling synergistic co-training. For instance, in a ``motion prediction'' VQA task, the model is prompted with image pairs $(I_t^i, I_{t+H}^i)$ and trained to predict the corresponding language-action $\hat{a}_{t:t+H}^{\text{lang.}}$ describing the observed displacement.

\subsection{Model Architecture and Training}
\label{subsec:architecture}

LAP is agnostic to the underlying VLA architecture and can, in principle, be integrated into any VLM-based policy by treating action prediction as a language modeling problem. However, auto-regressive generation of language-actions at inference time is slow and impractical for real-time robot control. To obtain an efficient system, we instantiate LAP as LAP-3B, which adopts a Mixture-of-Transformers architecture \cite{liang2024mixture} combining a LAP-trained VLM backbone with a lightweight flow-matching action expert for high-frequency motor execution, following $\pi_{0.5}$ \cite{intelligence2025pi05visionlanguageactionmodelopenworld}. This design enables a controlled comparison: \textsc{LAP-3B} differs from $\pi_{0.5}$ only in the action representation used to supervise the VLM (language-actions versus FAST tokens), isolating the downstream effect of the representation itself, which we evaluate in Sec.~\ref{sec:experiments}.

During training, the VLM backbone is optimized to predict structured language-actions using the cross-entropy loss in Eq.~\eqref{eq:language-action}, while the action expert predicts continuous action chunks $a_{t:t+H}$ via a flow-matching objective
\begin{equation}
\mathcal{L}_{\text{FM}}
=
\mathbb{E}_{(a_{t:t+H}, o_t, l) \sim \mathcal{D}}
\left[
\left\|
v_{\phi}(o_t, l, a_{t:t+H}, \tau) - \dot{x}_{\tau}
\right\|^2
\right],
\end{equation}
where $v_{\phi}$ denotes the velocity field parameterized by the action expert and $\dot{x}_{\tau}$ is the target flow corresponding to the ground-truth action chunk (see Appendix~\ref{app:implementation_details} for the full formulation). The overall training objective is
\begin{equation}
\mathcal{L}
=
\mathcal{L}_{\text{FM}}
+
\lambda\,\mathcal{L}_{\text{CE}}.
\end{equation}

Under this architecture, the VLM and action expert communicate solely through cross-attention. We apply causal masking to ensure that the raw action and the language-action do not attend to each other.
Following \cite{driess2025knowledgeinsulatingvisionlanguageactionmodels}, we further block gradients from the action expert propagating back into the VLM backbone. While not required by LAP, this knowledge insulation helps preserve pre-trained VLM representations and stabilizes joint training. At inference time, only the action expert is rolled out for control, enabling real-time execution at 25\,Hz on an NVIDIA RTX~4090 GPU.

\subsection{Implementation Details}
\label{sec:implementation_details}

\textsc{LAP-3B}'s VLM backbone is initialized as the PaliGemma-3B \cite{beyer2024paligemmaversatile3bvlm} model. We train on the Open X-Embodiment (OXE) dataset \cite{embodimentcollaboration2025openxembodimentroboticlearning} and MolmoAct dataset \cite{lee2025molmoactactionreasoningmodels} (see full data mixture details in Appendix~\ref{app:implementation_details}). We use a shuffle buffer size of 16 million samples to ensure near uniform mixing, which improves training stability.
For proprioception, we represent state $s_t$ using Cartesian end-effector pose: position, a continuous 6D rotation representation \cite{zhou2019continuity}, and a binary gripper state. Since open-source datasets use inconsistent rotation conventions, we canonicalize rotations to extrinsic XYZ before converting to the 6D representation. The proprioceptive state is discretized into 255 bins and appended to the prompt as text tokens, following \cite{intelligence2025pi05visionlanguageactionmodelopenworld}. Actions are represented as delta end-effector poses. We normalize states and actions using the 1st and 99th global quantiles computed over the full training dataset, and reuse the same statistics at inference on unseen embodiments.

Training uses a global batch size of 2048 across 64 TPU v6e chips for 15k gradient steps (\emph{approximately only 10 wall-clock hours}), corresponding to roughly 0.65 epoch over the full dataset, at which point the checkpoint already performs well in real-world experiments. Exponential moving average (EMA) updates begin after 5k steps.
Image inputs are resized to $224 \times 224$, with at most two images per sample. We use a fixed learning rate of $1 \times 10^{-4}$ with linear warmup over the first 5,000 steps.

\begin{figure*}[t]
    \centering
    \includegraphics[width=\linewidth]{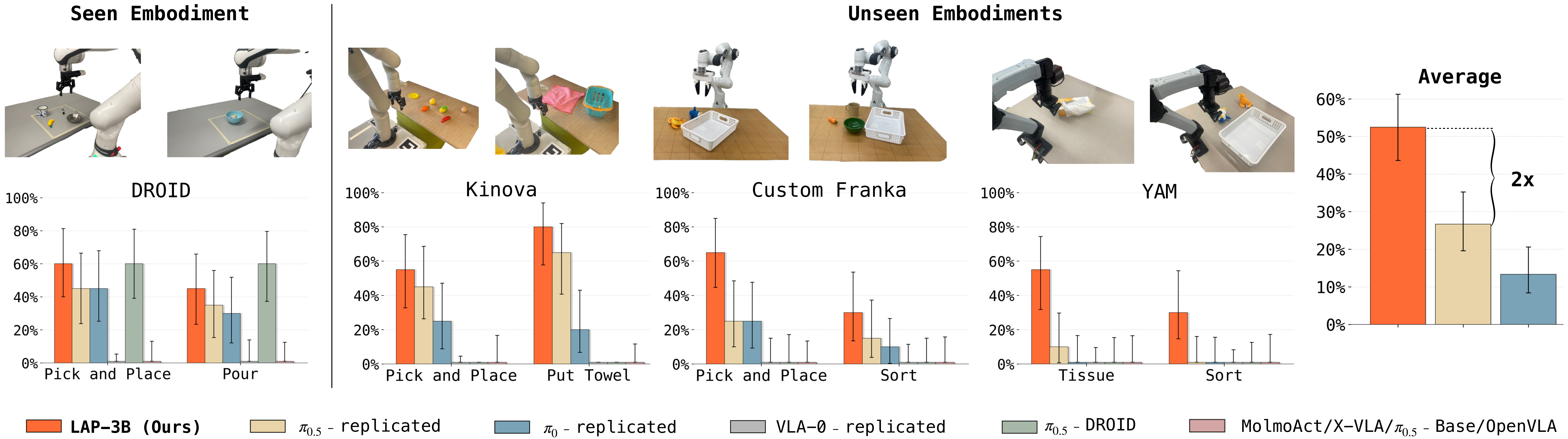}
    \caption{\textbf{Zero-shot cross-embodiment generalization performance.} LAP-3B achieves performance comparable to the $\pi_{0.5}$-DROID on the seen embodiment. Across three previously unseen embodiments and six real-world manipulation tasks, LAP-3B attains over \textbf{50\% average zero-shot success}, delivering approximately a $2\times$ improvement over the strongest baselines, while all open-sourced VLAs collapse to zero success rate. Error bars denote 95\% finite-sample–valid confidence intervals that control the Type-I error (miscoverage) probability, following~\cite{vincent2024generalizablebehaviorcloningpolicy}.
}
    \label{fig:zero-shot}
\end{figure*}

\section{Experiments}
\label{sec:experiments}

Our experiments aim to evaluate whether language-actions provide a superior action representation for VLA pre-training. In particular, we focus on \textbf{cross-embodiment generalization}, which remains a significant limitation of state-of-the-art VLAs. Concretely, we seek to answer:
{
\begin{enumerate}
    \item Does LAP enable zero-shot cross-embodiment transfer?
    \item Does LAP support efficient fine-tuning to new embodiments on challenging tasks?
    \item How does LAP’s language-action representation shape the learned representations and enable strong cross-embodiment generalization?
    \item What is the effect of co-training \textsc{LAP-3B} with VQA datasets?
    \item Does LAP scale favorably with model size?
\end{enumerate}
}
To address these questions, we evaluate LAP across \textbf{four robot embodiments}, \textbf{ten real-world manipulation tasks}, and the LIBERO simulation benchmark \cite{liu2023liberobenchmarkingknowledgetransfer}. We first test out-of-the-box performance on one seen and three unseen embodiments, then study fine-tuning efficiency on new robots and tasks. We investigate the impact of LAP's language-action representation on training dynamics and conclude with an analysis of co-training with Visual--Question--Answering (VQA) datasets.

\noindent\textbf{Baselines.}
We evaluate two categories of baselines below. All replicated baselines are carefully tuned (learning rate, batch size, loss weights, etc.) for fair comparison; full details are in Appendix~\ref{app:implementation_details}.

\paragraph{Open-Sourced VLAs.}
We evaluate several publicly released VLA checkpoints to assess whether existing methods already exhibit zero-shot cross-embodiment transfer:
(1) \textbf{\bm{$\pi_{0.5}$}-DROID} \cite{intelligence2025pi05visionlanguageactionmodelopenworld}, currently the strongest VLA on the DROID benchmark;
(2) \textbf{\bm{$\pi_{0.5}$}-Base} \cite{intelligence2025pi05visionlanguageactionmodelopenworld}, the strongest publicly available VLA base model to our knowledge;
(3) \textbf{X-VLA} \cite{zheng2025xvlasoftpromptedtransformerscalable}, which explicitly targets embodiment generalization;
(4) \textbf{MolmoAct} \cite{lee2025molmoactactionreasoningmodels}; and
(5) \textbf{OpenVLA} \cite{kim2024openvlaopensourcevisionlanguageactionmodel}.

\paragraph{Action-Representation Comparison.}
To isolate the impact of language-action supervision, we train controlled baselines with identical architectures and data mixtures, differing only in action representation:

\begin{enumerate}
    \item \textbf{$\bm{\pi_{0.5}}$-replicated}: supervises the VLM using FAST tokens \cite{pertsch2025fastefficientactiontokenization}. Since the FAST tokenizer is pre-trained on substantially more robot data than our model, this forms a particularly strong baseline.
    \item \textbf{$\bm{\pi_{0}}$-replicated}: removes action supervision from the VLM entirely and only supervises the action expert, measuring the benefit of language-action pre-training.
    \item \textbf{VLA-0-replicated}: supervises the VLM following \citet{goyal2025vla0buildingstateoftheartvlas}, where continuous actions are represented as digit tokens.
\end{enumerate}
Across all experiments, we conduct \textbf{over 1300 real-robot evaluation trials}.

\subsection{Zero-Shot Cross-Embodiment Generalization}
\label{subsec:zero_shot}

\noindent\textbf{Embodiments and Tasks.}
We evaluate on four robot embodiments (Fig.~\ref{fig:zero-shot}): the DROID setup \cite{khazatsky2025droidlargescaleinthewildrobot} which is included in the pre-training mixture, three unseen embodiments---a custom 7-DoF Franka Panda robot with a novel gripper and wrist-mounted camera configuration, the 6-DoF YAM robot, and the 7-DoF Kinova robot.

We evaluate five manipulation task types (Fig.~\ref{fig:zero-shot}, top row):
\begin{enumerate}
    \item \textbf{Pick and Place}: pick up an object and place it into a container, testing generalization across gripper geometry, camera placement, and robot appearance.
    \item \textbf{Sort}: sort all objects on the table into a basket, evaluating long-horizon manipulation.
    \item \textbf{Tissue}: pull tissue from a box and place it on the table, requiring adaptive grasping of deformable objects and height reasoning.
    \item \textbf{Put Towel}: pick up a flat towel and place it into a basket, requiring precise grasp localization and large vertical motion.
    \item \textbf{Pour}: pour contents from a bowl, testing rotational control and primitive sequencing.
\end{enumerate}

These tasks require full 6-DoF manipulation, e.g., rotating the gripper to avoid collisions during sorting, and probe complementary primitive skills.

For each embodiment, we evaluate on two of these tasks with 20 trials per task. We report mean success rates with 95\% confidence intervals following the method of \cite{vincent2024generalizablebehaviorcloningpolicy}, which provides statistically valid finite-sample coverage. Unlike heuristic standard deviations, these intervals explicitly control the Type-I error (miscoverage) probability. Success metric is binary: a trial is marked successful only if the full task is completed. Results are summarized in Fig.~\ref{fig:zero-shot}.

\noindent\textbf{Seen Embodiment Evaluation.} On the in-distribution DROID setup, we observe that nearly all open-sourced VLAs fail to achieve non-trivial task success, with the sole exception of $\pi_{0.5}-\text{DROID}$ \cite{intelligence2025pi05visionlanguageactionmodelopenworld}, which has been explicitly fine-tuned on the DROID dataset \cite{khazatsky2025droidlargescaleinthewildrobot}. This highlights a critical limitation of current VLAs: despite large-scale multi-embodiment pre-training, effective deployment even on seen embodiments typically requires embodiment-specific adaptation. 

For our replicated comparison baselines, $\pi_{0.5}$-replicated and $\pi_{0}$-replicated achieve moderate and comparable performance. However, $\text{VLA-0}$-replicated consistently collapses to near-zero actions. We find this failure mode stems from overfitting to trivial predictions, which manifests as predicting only small-magnitude action tokens. This issue may be less of a concern on smaller, cleaner benchmarks such as LIBERO \cite{liu2023liberobenchmarkingknowledgetransfer} but becomes detrimental at large scale, where noise and behavioral diversity are unavoidable.

Notably, \textsc{LAP-3B} consistently outperforms $\pi_{0.5}$-replicated and $\pi_{0}$-replicated by approximately 15 percentage points across tasks, despite sharing the same backbone architecture and training data; this highlights the superiority of the language-action representation. Moreover, \textsc{LAP-3B} achieves performance on par with $\pi_{0.5}$-DROID---without any embodiment-specific fine-tuning on the DROID dataset itself. This indicates that language-action supervision enables effective learning from large-scale, heterogeneous robot data, leading to strong in-distribution performance.

\noindent\textbf{Unseen Embodiment Evaluation.} We next evaluate zero-shot transfer to three previously unseen robot embodiments. Across all settings, \textsc{LAP-3B} demonstrates a substantial and consistent advantage, with an \emph{average success rate exceeding 50\%}, representing an approximate \textbf{2$\times$ improvement over the strongest baseline}. Importantly, \textsc{LAP-3B} is the only model to achieve consistently high performance across all novel embodiments.

In contrast, all open-sourced VLAs, including $\pi_{0.5}$-DROID, which performed strongly in-distribution, fail to generate meaningful behavior on unseen robots. This sharp collapse suggests that existing VLA training recipes remain tightly coupled to embodiment-specific control conventions, preventing transfer even when visual and task semantics remain similar.

By contrast, \textsc{LAP-3B} exhibits both \emph{semantically grounded movements} and \emph{fine-grained spatial control} across embodiments. We hypothesize that the language-action interface provides a shared, embodiment-agnostic abstraction that better preserves the VLM’s generalizable visual-semantic representations while enabling precise execution under novel robot kinematics. This decoupling allows \textsc{LAP-3B} to retain coherent task intent while avoiding the embodiment-specific overfitting that hampers existing VLAs.

\begin{figure*}[t]
    \centering
    \includegraphics[width=\linewidth]{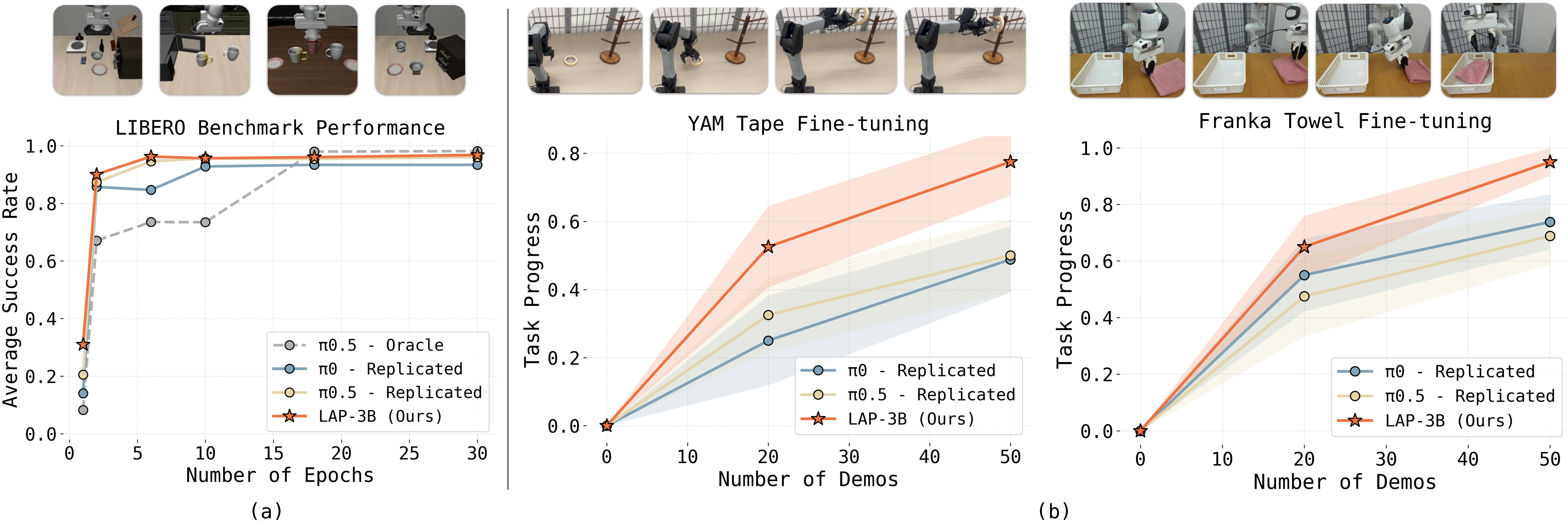}
    \caption{\textbf{Fine-tuning efficiency in simulation (LIBERO) and real-world manipulation tasks}. Across both domains, \textsc{LAP-3B} adapts substantially faster than baseline policies, reaching high performance with significantly fewer epochs and demonstrations. In simulation, \textsc{LAP-3B} converges to near-optimal success within a fraction of the training steps required by baselines. On real robots, \textsc{LAP-3B} achieves comparable task performance using approximately \textbf{$\bm{2.5\times}$ fewer demonstrations}, demonstrating substantially improved data and compute efficiency when transferring to new embodiments.}
    \label{fig:libero-fine-tuning}
\end{figure*}
\vspace{2pt}

\subsection{Fine-Tuning Efficiency on New Embodiments}
\label{sec:finetuing}

While \textsc{LAP-3B} exhibits non-trivial zero-shot performance on novel embodiments for simple tasks, more challenging manipulation scenarios still require embodiment-specific adaptation at the current training scale. We therefore investigate whether LAP improves \emph{fine-tuning efficiency} along two axes: \emph{data efficiency}---requiring fewer demonstrations to reach a given performance level, and \emph{compute efficiency}---converging with fewer gradient steps.

\noindent\textbf{Simulation Fine-Tuning Experiments.}
We first fine-tune \textsc{LAP-3B} on the LIBERO simulation benchmark \cite{liu2023liberobenchmarkingknowledgetransfer} and compare against $\pi_{0.5}$-replicated and $\pi_0$-replicated baselines trained on the same pre-training mixture. We additionally report an oracle baseline, $\pi_{0.5}$-Oracle, initialized from the original $\pi_{0.5}$ weights pre-trained on substantially larger datasets.

As shown in Fig.~\ref{fig:libero-fine-tuning}(a), \textsc{LAP-3B} converges substantially faster than all baselines. With only a single epoch of fine-tuning, \textsc{LAP-3B} already reaches 78\% success, and achieves near-maximum performance (96.8\%) within six epochs---significantly fewer updates than both replicated baselines and the oracle model \cite{intelligence2025pi05visionlanguageactionmodelopenworld}. While $\pi_{0.5}$-Oracle eventually attains slightly higher asymptotic performance after extended training, \textsc{LAP-3B} demonstrates markedly superior compute efficiency. These results indicate that language-action pre-training yields a more transferable initialization that adapts rapidly to new embodiments. We summarize LIBERO benchmark results for \textsc{LAP-3B} and baseline methods in Appendix~\ref{app:additional_results}.

\noindent\textbf{Real-World Fine-Tuning Experiments.}
We next fine-tune on two challenging real-robot tasks (Fig.~\ref{fig:libero-fine-tuning} top row, right):
\begin{itemize}
    \item \textbf{Hang Tape on Rack}: pick up the tape by its edge and hang it on the top rack, requiring dexterity and precision. The task is evaluated on the YAM robot.
    \item \textbf{Fold Towel and Place in Basket}: fold the towel in half from left to right, then place the folded towel into a basket, requiring both fine-grained dexterity and long-horizon manipulation. The task is evaluated on the Custom Franka robot.
\end{itemize}
Results are shown in Fig.~\ref{fig:libero-fine-tuning}. We evaluate task progress metrics over 20 trials per data point (see Appendix~\ref{app:evaluation_details} for details).

Across both tasks, \textsc{LAP-3B} achieves strong performance with substantially fewer demonstrations. On YAM, \textsc{LAP-3B} reaches approximately 50\% task progress using only 20 demonstrations---roughly \textbf{$\bm{2.5\times}$ demos fewer than baselines}. On Franka, \textsc{LAP-3B} consistently outperforms $\pi_{0.5}$-replicated and $\pi_0$-replicated across all data regimes. Together, these results demonstrate that language-action pre-training produces a more adaptable VLA, improving both sample efficiency and fine-tuning speed when transferring to new embodiments and tasks.

\begin{figure*}[t]
    \centering
    \includegraphics[width=\linewidth]{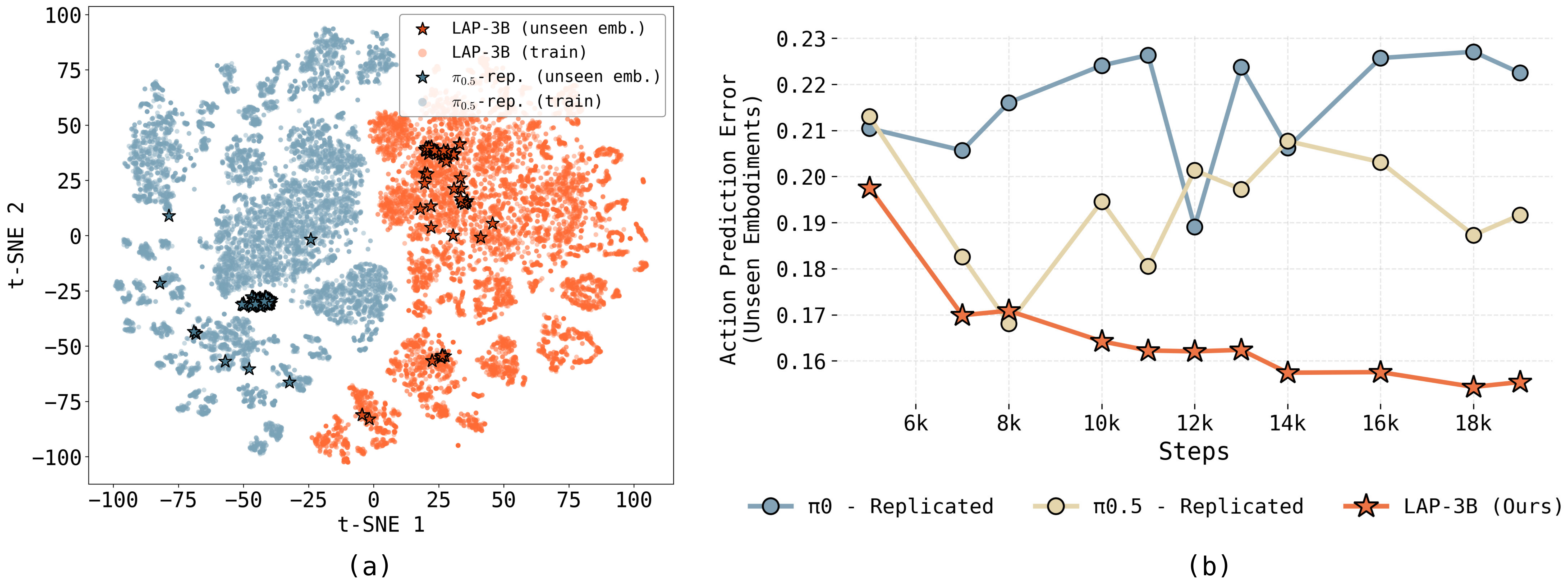}
    \caption{
    \textbf{(a) T-SNE visualizations of learned embodiment representations} for \textsc{LAP-3B} and $\pi_{0.5}$-replicated. \textsc{LAP-3B} exhibits substantial overlap between training and unseen embodiments, whereas $\pi_{0.5}$-replicated shows limited alignment, indicating that \textsc{LAP-3B} learns more transferable, embodiment-agnostic control representations.
    \textbf{(b) Action prediction error on unseen embodiments during pre-training.} \textsc{LAP-3B} achieves consistently lower action prediction error on held-out unseen embodiments throughout training, compared to $\pi_{0.5}$-replicated and $\pi_{0}$-replicated baselines. This indicates that language-action supervision enables the model to learn control representations that generalize across embodiments, allowing more accurate action prediction on novel robots as well as smoother training dynamics.
    }
    \label{fig:analysis}
\end{figure*}

\subsection{Analyzing LAP’s Cross-Embodiment Generalization}

\noindent\textbf{T-SNE visualization of embodiment representations.}
We visualize learned embodiment features using t-SNE in Fig.~\ref{fig:analysis}(a) by embedding the average prelogits of action tokens, which capture internal control representations across robot embodiments. \textsc{LAP-3B} exhibits substantially improved alignment between unseen embodiments and training data, with features from unseen embodiments overlapping closely with those from training embodiments rather than forming separate clusters. In this representation space, stronger overlap between training and unseen embodiments indicates that the model has learned more transferable, embodiment-agnostic representations, helping explain why language-action pre-training enables strong zero-shot embodiment transfer.

\noindent\textbf{Low action prediction error on unseen embodiments.}
We next evaluate the quality of representations learned during pre-training by measuring action prediction error on held-out datasets from \emph{unseen} robot embodiments described in Sec.~\ref{sec:finetuing}. Prediction error is measured as the $\ell_2$ distance between predicted and ground-truth actions.

As shown in Fig.~\ref{fig:analysis}(b), \textsc{LAP-3B} consistently achieves lower action prediction error on unseen embodiments than baselines trained with alternative action representations, indicating that language-action supervision leads to more generalizable control representations across embodiment shifts. In addition, the prediction error decreases smoothly and monotonically over training, suggesting that LAP enables more stable training dynamics.

\begin{wraptable}{r}{0.5\columnwidth}
\vspace{-12pt}
\centering
\scriptsize
\setlength{\tabcolsep}{2pt}
\renewcommand{\arraystretch}{0.95}
\caption{Best prediction errors on held-out datasets from \emph{unseen robot embodiments} for \textsc{LAP-3B} and baseline methods. Best per row is bolded.}
\label{tab:val_losses}
\begin{tabular}{lccc}
\toprule
 & \textbf{\textsc{LAP-3B}} & $\pi_{0.5}$-replicated & $\pi_{0}$-replicated \\
\midrule
Prediction Error (Unseen) & \textbf{0.151} & 0.168 & 0.189 \\
Validation Loss (Unseen) & \textbf{0.049} & 0.051 & 0.052 \\
\bottomrule
\end{tabular}
\vspace{-12pt}
\end{wraptable}

We additionally report the best validation loss of the (flow-matching) action expert in Table~\ref{tab:val_losses}. Taken together, these results suggest that language-action pre-training does not merely stabilize optimization, but fundamentally improves how control representations transfer across embodiments—allowing \textsc{LAP-3B} to predict more accurate actions on unseen robots and explaining both its strong zero-shot performance and superior fine-tuning efficiency.

\subsection{Co-Training with Vision-Language Datasets}
\begin{wrapfigure}{r}{0.3\linewidth}
   \centering
   \vspace{-0.7cm}
  \includegraphics[width=\linewidth]{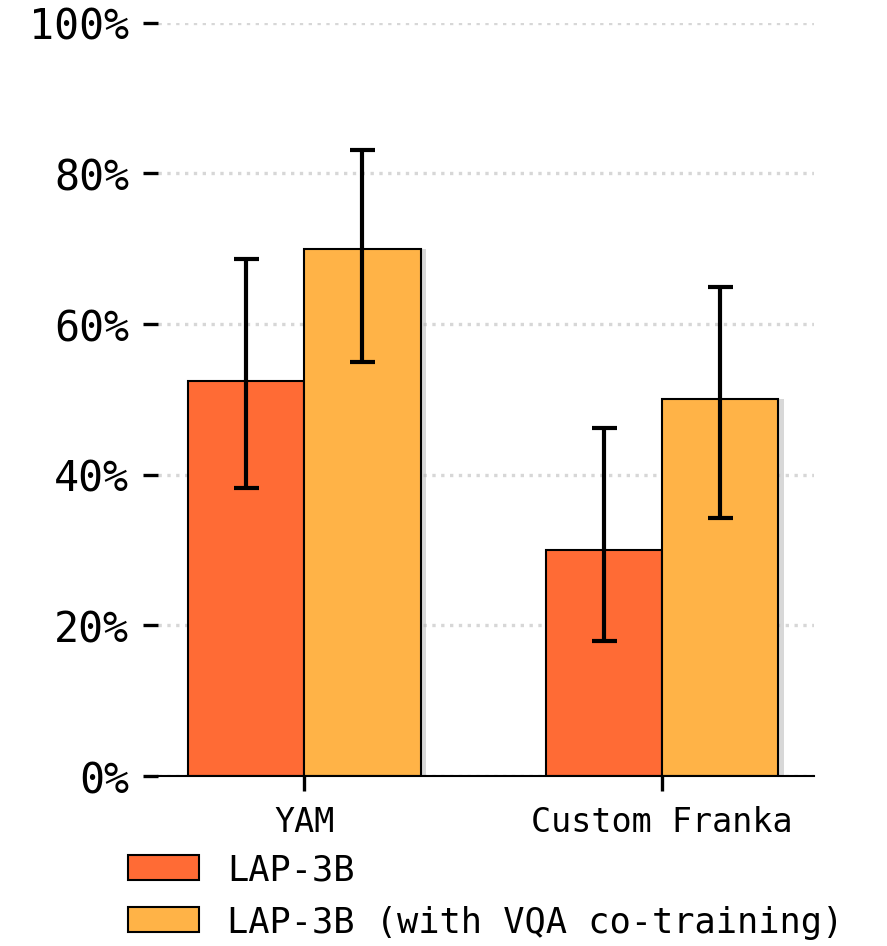}
  \caption{\textbf{VQA co-training} with a motion-prediction task improves cross-embodiment performance.}
  \label{fig:cotrain}
  \vspace{-0.5cm}
\end{wrapfigure}
We further investigate whether co-training with additional Visual-Question-Answering (VQA) tasks benefits \textsc{LAP-3B}.
We consider a motion-prediction task in which the model is prompted with two consecutive images and asked to predict the corresponding language-action describing the change between frames, analogous to an ``inverse dynamics'' prediction (Fig.~\ref{fig:teaser}, bottom left).
This co-training objective naturally bridges standard VQA-style supervision and our language-action prediction by sharing a unified, language-based output format.

We evaluate the resulting models on two embodiments, Custom Franka and YAM, with results shown in Fig.~\ref{fig:cotrain}.
Incorporating this motion-prediction VQA task leads to more precise action generation and improved spatial generalization, yielding higher success rates across embodiments.
In addition, co-training produces checkpoints that are easier to adapt to new tasks, enabling faster fine-tuning with fewer gradient steps and achieving higher final convergence performance (see Appendix~\ref{app:additional_results}).

\subsection{Scaling with Model Size}
\begin{wrapfigure}{r}{0.5\linewidth}
    \centering
    \vspace{-0.4cm}
    \includegraphics[width=\linewidth]{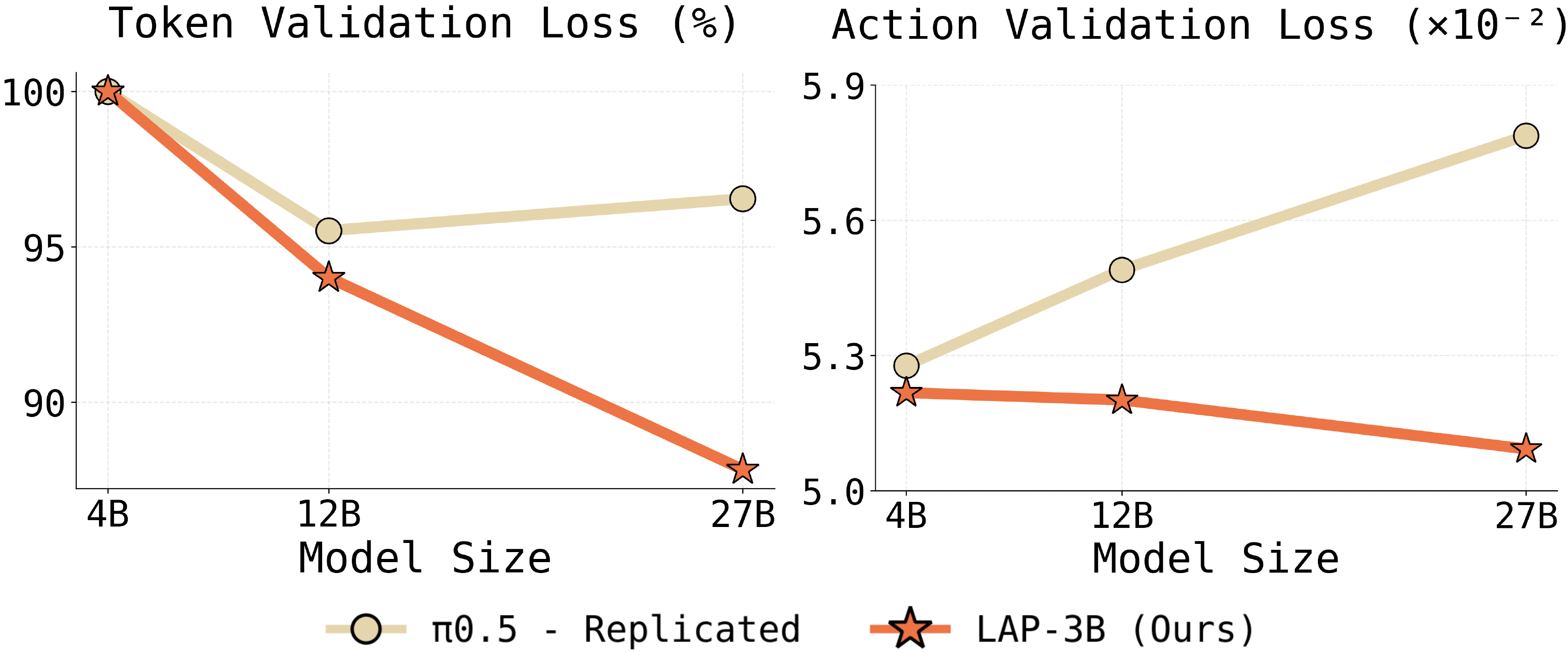}
    \caption{\textbf{Model scaling behavior of LAP compared to $\bm{\pi_{0.5}}$-replicated.}
    Left: Token validation loss, reported as percentage drop relative to the 4B model.
    Right: Continuous action validation loss of the diffusion-based action expert.
    \textsc{LAP-3B} improves consistently with scale, whereas the baseline saturates and degrades.}
    \label{fig:model-scaling}
\end{wrapfigure}

We next investigate whether LAP benefits from increased model capacity, and whether its advantages persist in the large-model regime.

Using Gemma3 \cite{gemmateam2025gemma3technicalreport} as the VLM backbone, we train LAP models at 4B, 12B, and 27B parameter scales, and compare them against corresponding $\pi_{0.5}$-replicated variants, which represent the current state of the art in VLAs.

We evaluate two metrics on a held-out split comprising 2.5\% of the full data mixture: (i) token validation loss of the VLM backbone, and (ii) validation loss of the flow-matching-based action expert. For token validation loss, we report the percentage reduction relative to the 4B model, as absolute loss values are not directly comparable across different action representations. 

As shown in Fig.~\ref{fig:model-scaling}, LAP consistently attains lower validation loss across all model sizes and exhibits favorable scaling behavior, with performance improving monotonically as model capacity increases. In contrast, the $\pi_{0.5}$-replicated baseline shows early saturation and even degradation at larger scales. These results indicate that language-action supervision remains effective in the large-model regime and increasingly benefits from additional representational capacity.

\section{Conclusions and Discussions}
We introduce Language-Action Pre-training (LAP), a scalable VLA pre-training recipe that predicts actions in natural language, yielding representations that transfer effectively across robot embodiments. We further present \textsc{LAP-3B}, which achieves substantial zero-shot generalization on real robots—delivering roughly a 30\% absolute improvement over prior action representations and, to our knowledge, the first substantial zero-shot success rate on novel embodiments without fine-tuning. \textsc{LAP-3B} also exhibits stable training, favorable scaling, and strong fine-tuning efficiency.

\noindent\textbf{Broader implications.}
LAP opens the possibility of training VLMs \emph{explicitly for robotics}, enabling robot (and potentially non-robot) interaction data to be seamlessly mixed with standard VLM corpora to learn models that reason, communicate, and act within a unified representation space.

\noindent\textbf{Limitations and future work.}
Despite these promising results, several limitations remain. While LAP can be applicable to substantially more heterogeneous robot systems and data sources, this paper focuses only on zero-shot transfer across single-arm manipulators. For example, LAP can be easily adapted to bimanual robots, which actions can be represented through coordinated dual-arm descriptions (e.g., Left Arm: move forward 5 cm; Right Arm: rotate clockwise 20 degrees). Another key advantage of language-actions is that they naturally operate at varying levels of precision and are more
tolerant to less accurate action labels, enabling the use of data such as human pose tracking, UMI, or internet video where precise low-level control signals are difficult to obtain. Systematically investigating these directions at scale remains an important avenue for future work.

Second, although \textsc{LAP-3B} adapts efficiently to moderately dexterous tasks such as \emph{Fold Towel and Put in Basket} and \emph{Hang Tape on Rack}, we have not yet evaluated regimes requiring substantially higher control frequency or extreme precision (e.g., fast reactive control, or fine-grained deformable object manipulation). Extending language-action supervision to these settings—potentially through hierarchical or multi-scale action representations—remains an open challenge.
\section*{Acknowledgments}

We thank Google’s TPU Research Cloud (TRC) for providing access to Cloud TPUs. We also thank Maria Attarian, Samuel Bateman and Julian Mendes for their valuable feedback on this project.
The authors were partially supported by the NSF CAREER Awards \#2044149 and \#2107048, 
the Office of Naval Research under Grant N00014-23-1-2148, the Army Research Laboratory under DCIST CRA W911NF-17-2-0181, a Sloan Fellowship, and Gemini Academic Program. Asher J.~Hancock was supported by the NSF Graduate Research Fellowship (DGE-2146755).

\clearpage

\bibliographystyle{unsrtnat}
\bibliography{references}

\clearpage
\beginappendix{
    \section{Additional Qualitative and Benchmark Results}
\label{app:additional_results}

\subsection{Qualitative Analysis for Zero-Shot Cross-Embodiment Experiments}
We present additional qualitative comparisons of all baseline policies under zero-shot embodiment transfer, complementing the results in Section~\ref{subsec:zero_shot}. 
OpenVLA~\cite{kim2024openvlaopensourcevisionlanguageactionmodel}, MolmoAct~\cite{lee2025molmoactactionreasoningmodels}, and $\pi_{0.5}$-Base \cite{intelligence2025pi05visionlanguageactionmodelopenworld} consistently fail to produce meaningful motions across all robot platforms. 
X-VLA~\cite{zheng2025xvlasoftpromptedtransformerscalable} generates smooth trajectories and occasionally approaches target objects on the DROID and Custom Franka setups; however, it never achieves accurate gripper–object alignment required for successful manipulation.

The $\pi_{0}$-replicated baseline exhibits highly imprecise end-effector control. For instance, in the \emph{Tissue} task on the YAM robot, it frequently overshoots the target and grasps the tissue box rather than the tissue itself. In contrast, $\pi_{0.5}$-replicated generally produces smoother and more directionally reasonable motions, but still lacks the spatial precision required for reliable grasping and successful task completion. Across the Custom Franka and Kinova robots, both baselines similarly exhibit imprecise motions, suggesting limited ability to infer accurate control actions under embodiment shift. These failure modes are consistent across all evaluated embodiments. An example rollout is shown in Fig.~\ref{fig:baseline-rollout-visualization}.

\begin{figure*}[h]
    \centering
    \includegraphics[width=\linewidth]{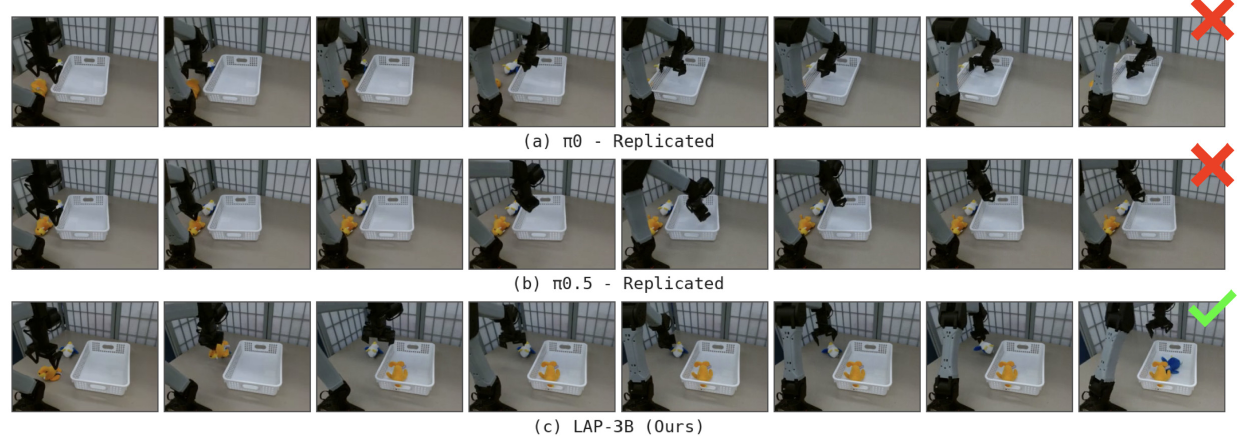}
   \caption{\textbf{Example baseline rollout failures on the Sort task (YAM robot).} 
    Both $\pi_{0}$-replicated and $\pi_{0.5}$-replicated exhibit imprecise end-effector control, moving directly toward the basket without first grasping any objects. 
    In contrast, LAP-3B successfully completes the task under similar initial conditions. 
    Open-sourced VLA baselines are omitted as they do not generate meaningful or safe control actions in this setting.}
    \label{fig:baseline-rollout-visualization}
\end{figure*}
\vspace{-5pt}


\begin{figure*}[b]
    \centering
    \includegraphics[width=\linewidth]{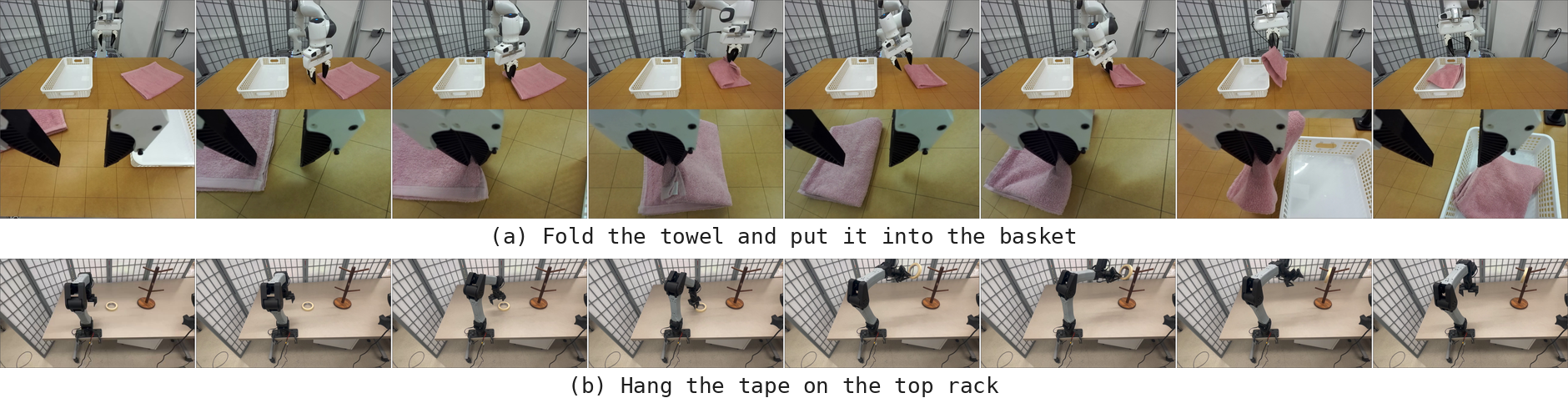}
    \caption{\textbf{Fine-tuning evaluation rollout visualizations.} Example observations from the camera views used for policy evaluation across different robot embodiments and tasks, shown from left to right.}
    \label{fig:finetuning-rollout-visualization}
\end{figure*}
\vspace{2pt}

\begin{table}[!h]
\centering
\footnotesize
\setlength{\tabcolsep}{4pt}
\caption{LIBERO benchmark results comparing LAP-3B with other state-of-the-art VLAs. Each data point corresponds to an average over 500 trials. Best per column is bolded. Rank is based on Avg. performance (higher is better).}
\begin{tabular}{l c c c c c c}
\toprule
Method & Spatial & Object & Goal & LIBERO-10 & Avg. & Rank \\
\midrule

TraceVLA \cite{zheng2025tracevlavisualtraceprompting} 
& 84.6 & 85.2 & 75.1 & 54.1 & 74.8 & 13 \\

X-VLA \cite{zheng2025xvlasoftpromptedtransformerscalable} 
& 98.2 & 98.6 & 97.8 & \textbf{97.6} & \textbf{98.1} & 1 \\

Octo \cite{octomodelteam2024octoopensourcegeneralistrobot} 
& 78.9 & 85.7 & 84.6 & 51.1 & 75.1 & 12 \\

UniAct \cite{zheng2025universalactionsenhancedembodied} 
& 77.0 & 87.0 & 77.0 & 70.0 & 77.8 & 11 \\

GR00T-N1 \cite{nvidia2025gr00tn1openfoundation} 
& 94.4 & 97.6 & 93.0 & 90.6 & 93.9 & 8 \\

OpenVLA \cite{kim2024openvlaopensourcevisionlanguageactionmodel} 
& 84.7 & 88.4 & 79.2 & 53.7 & 76.5 & 10 \\

OpenVLA-OFT \cite{kim2025finetuningvisionlanguageactionmodelsoptimizing} 
& 97.6 & 98.4 & 97.9 & 94.5 & 97.1 & 3 \\

UniVLA \cite{bu2025univlalearningacttaskcentric} 
& 95.4 & 98.8 & 93.6 & 94.0 & 95.4 & 6 \\

$\pi_0$ \cite{black2024pi0visionlanguageactionflowmodel} 
& 96.8 & 98.8 & 95.8 & 85.2 & 94.2 & 7 \\

$\pi_{0.5}$ \cite{intelligence2025pi05visionlanguageactionmodelopenworld} 
& 98.8 & 98.2 & 98.0 & 92.4 & 96.9 & 4 \\

VLA-0 \cite{goyal2025vla0buildingstateoftheartvlas} 
& 97.0 & 97.8 & 96.2 & 87.6 & 94.7 & 9 \\





\midrule
\textbf{\textsc{LAP-3B}} 
& 98.2 & \textbf{99.0} & \textbf{98.8} & 91.2 & 96.8 & 5 \\

\textbf{\textsc{LAP-3B}+ VQA Co.} 
& \textbf{99.0} & \textbf{99.0} & 97.2 & 93.4 & 97.2 & 2 \\

\bottomrule
\end{tabular}
\label{tab:libero_full}
\end{table}

\subsection{Full Results for the LIBERO Benchmark}
\begin{wrapfigure}{r}{0.4\textwidth}
    \centering
    \vspace{-20pt}
    \includegraphics[width=0.4\textwidth]{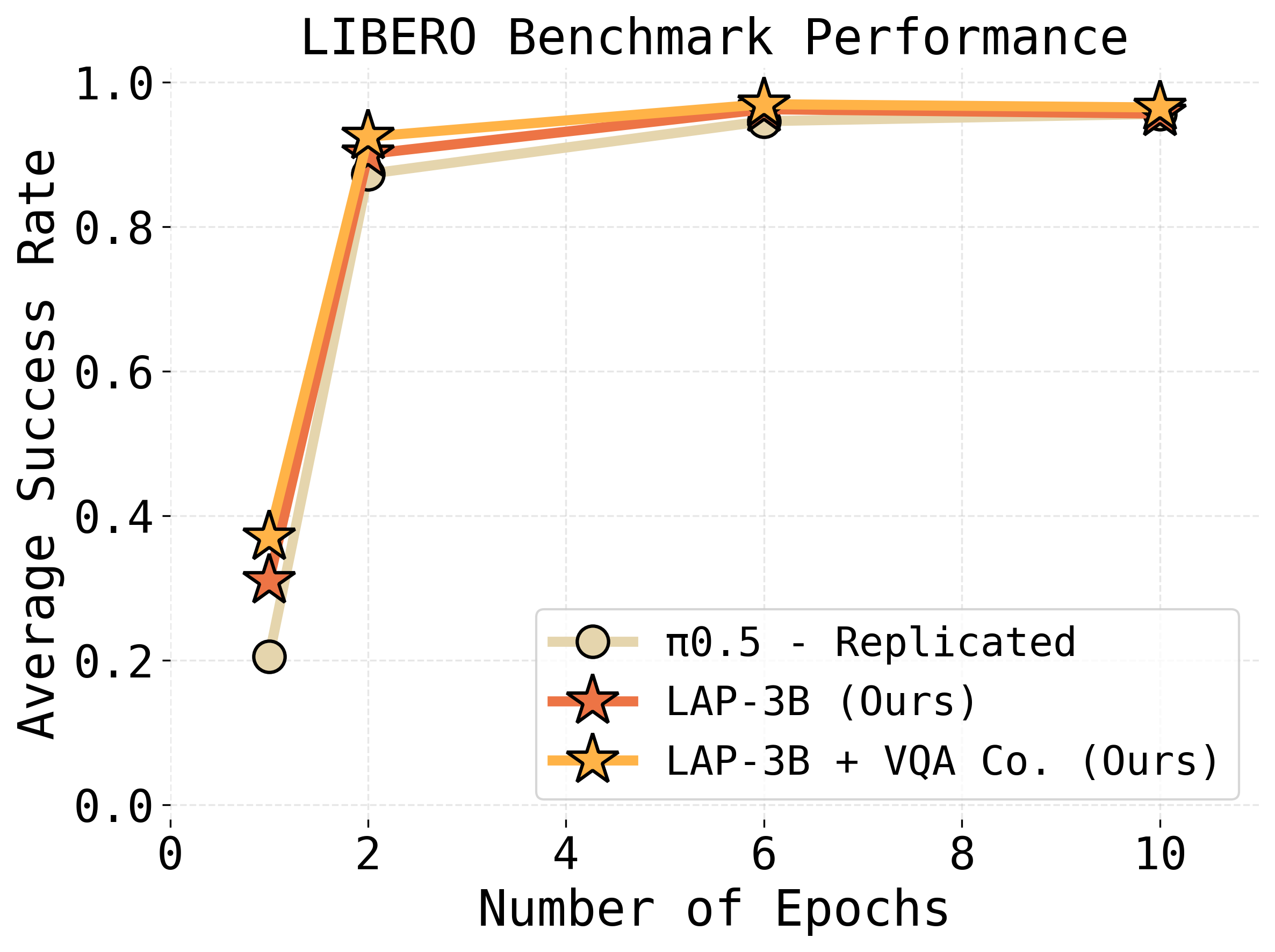}
    \captionsetup{skip=2pt}
    \caption{\textbf{Compute efficiency on LIBERO.} Co-training with VQA data significantly accelerates convergence, enabling \textsc{LAP-3B}+VQA Co-training to reach high success rates in fewer gradient steps. For clarity, we report only the strongest baseline ($\pi_{0.5}$-replicated).}
    \vspace{-1cm}
    \label{fig:libero-cotraining}
\end{wrapfigure}
Table~\ref{tab:libero_full} reports the complete LIBERO benchmark results, including additional open-source VLA baselines not shown in the main paper, and further confirms the consistent performance gains of \textsc{LAP-3B} across task families and all three task suites. \textsc{LAP-3B}+VQA Co. denotes the policy co-trained with VQA data as described in Section~\ref{sec:experiments}. Its improved performance over \textsc{LAP-3B} alone further underscores the effectiveness of VQA co-training. Moreover, the co-trained checkpoint substantially improves fine-tuning efficiency on LIBERO, achieving stronger performance with the same gradient steps, as shown in Fig.~\ref{fig:libero-cotraining}.

\subsection{Real-Robot Rollouts}
We further provide extended qualitative rollout visualizations across all embodiments and tasks for fine-tuning evaluation in Fig.\ref{fig:finetuning-rollout-visualization} and zero-shot evaluation in Fig.~\ref{fig:zeroshot-rollout-visualization}. 
Each panel corresponds to the camera views used during policy execution.

\section{Implementation Details}
\label{app:implementation_details}

\subsection{Robot Platforms} 
We evaluate \textsc{LAP-3B} across four robot embodiments that differ in kinematics, sensing configurations, and control interfaces:
\begin{itemize}
\item \textbf{DROID}: a seen embodiment during pre-training, comprising a 7-DoF arm with both external and wrist-mounted cameras.
\item \textbf{Custom Franka}: an unseen 7-DoF arm with external and wrist-mounted cameras, featuring a novel gripper and wrist camera mounting that differ from Franka robots observed during training.
\item \textbf{YAM}: an unseen 6-DoF arm with a single external camera.
\item \textbf{Kinova}: an unseen 7-DoF arm with external and wrist-mounted cameras.
\end{itemize}

For fine-tuning, we use joint-space actions on YAM to enable more precise low-level control, while all other robots are controlled using end-effector pose actions.

\subsection{Training Hyperparameters} 
\begin{wraptable}{r}{0.4\textwidth}
    \centering
    \small
    \vspace{-1.2cm}
    \caption{Training Hyperparameters for LAP-3B.}
    \label{tab:hyperparams}
    \renewcommand{\arraystretch}{1.15}
    \begin{tabular}{l c}
    \toprule
    \textbf{Hyperparameter} & \textbf{Value} \\
    \midrule
    Batch size & 2048 \\
    Learning rate & $1 \times 10^{-4}$ \\
    Warmup steps & 5000 \\
    EMA start step & 5000 \\
    EMA decay & 0.999 \\
    Action horizon & 16 \\
    Adam $(\beta_1, \beta_2)$ & $(0.9,\ 0.95)$ \\
    Gradient clipping norm & 1.0 \\
    Weight decay & $1 \times 10^{-4}$ \\
    \bottomrule
    \end{tabular}
    \vspace{-8pt}
\end{wraptable}
The overall training objective is:
\[
\mathcal{L} = \mathcal{L}_{\text{flow}} + \lambda \mathcal{L}_{\text{CE}}.
\]

We use $\lambda=0.8$ during pre-training and $\lambda=0.4$ during fine-tuning. Other hyper-parameters is shown in Table.\ref{tab:hyperparams}. We set $\lambda<1$ during pre-training to slightly downweight the language-action loss, as it converges substantially faster than the diffusion-based action expert. This balancing prevents the faster language-action objective from dominating optimization and encourages both components to learn at comparable rates.

\subsection{Baseline Implementations}
For all replicated baselines ($\pi_{0.5}$-replicated, $\pi_{0}$-replicated, and VLA-0-replicated), we carefully tune hyperparameters and select the best-performing checkpoints based on preliminary real-robot validation runs. 
Across all baselines, we sweep the learning rate over $\{1\times10^{-4}, 5\times10^{-5}\}$ and the batch size over $\{1024, 2048\}$. 
For $\pi_{0.5}$-replicated, we additionally sweep the cross entropy loss weight of the FAST tokens over $\{0.1, 0.4, 0.6, 0.8\}$ and the action horizon over $\{16, 32\}$.


Unless otherwise specified, the final hyperparameters used for all replicated baselines are a batch size of $2048$, learning rate of $1\times10^{-4}$, action horizon of $16$, and loss weight of $\lambda=0.6$.

Beyond these explicitly tuned parameters, all replicated baselines share the same design choices as our method, including identical network architectures, data mixtures, sampling strategies, and optimization settings. 
All remaining hyper-parameters follow those listed in Table~\ref{tab:hyperparams}.

For $\pi_{0}$-replicated, we directly follow the original OpenPI codebase implementation \cite{black2024pi0visionlanguageactionflowmodel}. 

For $\pi_{0.5}$-replicated, we extend the original implementation to include supervision of the VLM backbone using FAST action tokens and Knowledge Insulation \cite{driess2025knowledgeinsulatingvisionlanguageactionmodels} (stop gradient from the action expert to the backbone), which is not provided in the released code. 

For VLA-0-replicated, we reimplement the method strictly following the specifications in the original paper~\cite{goyal2025vla0buildingstateoftheartvlas}, including masked action augmentation, the recommended action resolution, and prompt format. We additionally perform careful hyperparameter tuning beyond what is reported in the paper, particularly for learning rate and batch size. With these adjustments, the replicated model achieves comparable performance on the LIBERO benchmark; however, it does not scale to large-scale datasets, as discussed in Section.\ref{subsec:zero_shot}.

\subsection{Prompt and Language-Action Format}
For both our model and all baselines, the prompt follow a structured template:

\begin{tcolorbox}[promptbox]
Task: \texttt{<instruction>}, predict the robot's action in the \texttt{<base frame>} or \texttt{<end-effector frame>}; State: \texttt{$s_1$\ $s_2$\ \ldots\ $s_D$}; Answer:
\end{tcolorbox}

An example example is:

\begin{tcolorbox}[promptbox]
Task: Put the marker into the cup, predict the robot's action in the base frame; State: 20 121 34 144 112 45 235 44 21 255; Answer:
\end{tcolorbox}

Language-actions are generated as structured clauses in fixed order:

\begin{tcolorbox}[promptbox]
\texttt{move <direction> <k> cm; tilt <direction> <k> degrees; rotate <cw/ccw> <k> degrees; open/close gripper}
\end{tcolorbox}

Translational magnitudes are discretized into integer centimeters and rotational magnitudes in integer degrees, with zero-valued movements omitted to suppress near-zero control noise.

\subsection{Motion Prediction VQA Data Generation}
Analogous to language–action data curation, motion prediction VQA pairs are generated automatically without manual annotation. Each training instance is constructed using the following prompt template:
\begin{tcolorbox}[promptbox]
Task: \texttt{question}; State: \texttt{$s_1$\ $s_2$\ \ldots\ $s_D$}; Answer:
\end{tcolorbox}
\noindent where the \texttt{question} ask the model to describe the temporal relationship between two observations (e.g., \emph{“What movement did the robot make from the first image to the second in the robot base frame?”}). We randomly sample semantically equivalent question phrasings to increase prompt diversity and robustness. The corresponding answer is given by the ground-truth language–action describing the robot motion between the two frames.

\subsection{Image Pre-processing and Augmentation}
All input images are first resized (with padding) to a fixed resolution before further processing. During training, we apply augmentation to images from all camera views. Images are normalized to $[0,1]$ and transformed using a composed pipeline consisting of a random crop to $95\%$ of the target resolution followed by resizing back to the original size, a random in-plane rotation sampled uniformly from $[-5^\circ, 5^\circ]$, and color jitter with brightness, contrast, and saturation parameters each set to $0.2$ (corresponding to multiplicative factors sampled uniformly from $[0.8, 1.2]$). The augmented images are then rescaled back to $[-1,1]$ for model input. Each image in the batch is augmented using an independent random seed. For VQA samples, augmentation is skipped entirely to preserve the original visual input.

\subsection{Attention Mask} 
\begin{wrapfigure}{r}{0.44\linewidth}
    \centering
    \vspace{-1.7cm}
    \includegraphics[width=\linewidth]{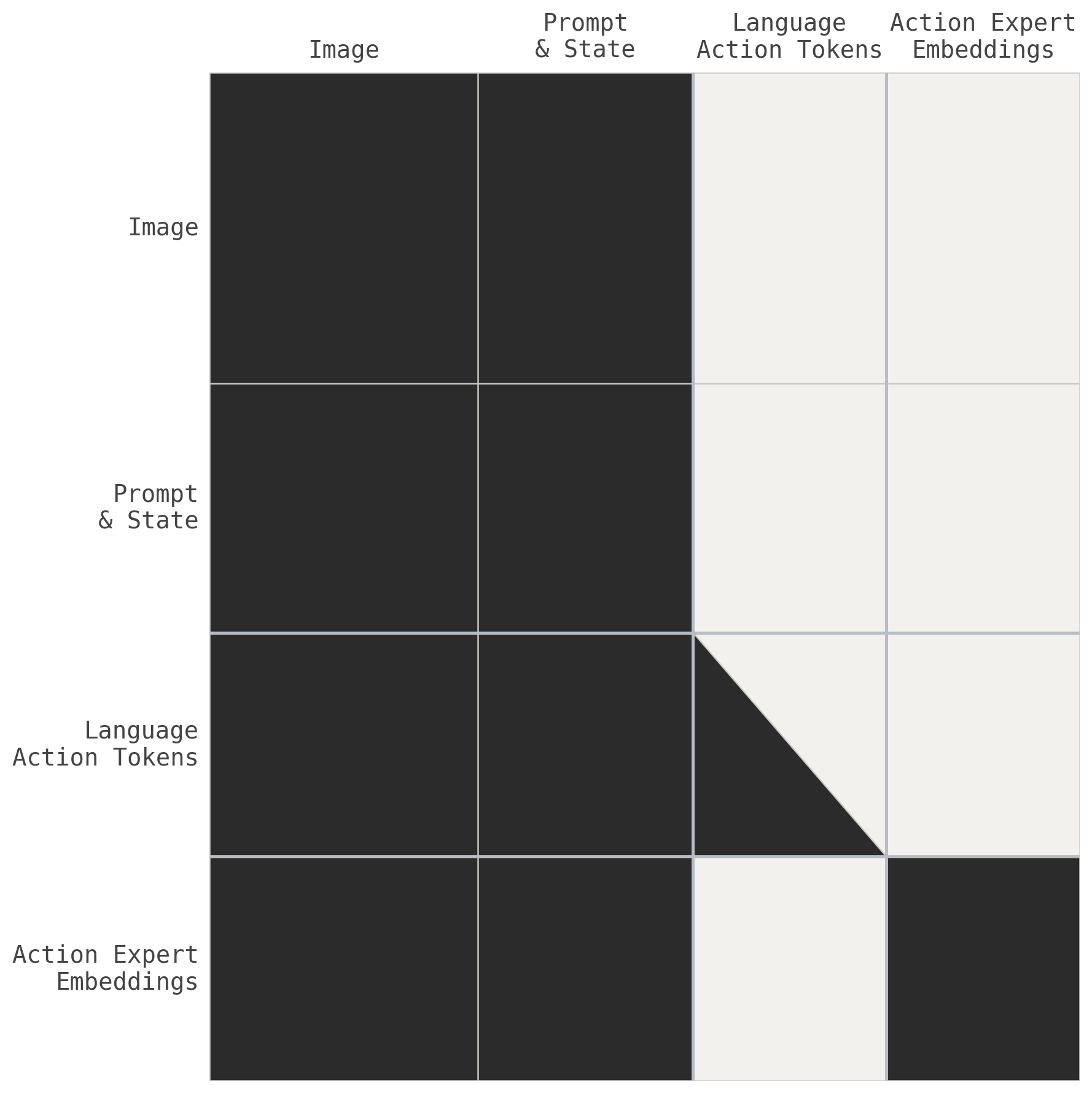}
    \captionsetup{skip=3pt}
    \caption{Attention mask visualization for LAP-3B.}
    \label{fig:attention-mask}
    \vspace{-6pt}
\end{wrapfigure}
We visualize the attention mask of \textsc{LAP-3B} in Fig.~\ref{fig:attention-mask}. Prefix tokens—including image, prompt, and state—use bidirectional attention. Language-action tokens attend causally to one another while having full attention to all prefix tokens. The action expert uses full self-attention and fully attends to prefix tokens, but does not attend to language-action tokens.

\subsection{Flow-Matching Objective}
Given a ground-truth action chunk $a$ and base noise $z \sim \mathcal{N}(0,I)$, we construct an interpolated state
\[
x_\tau = (1-\tau)z + \tau a, \quad \tau \sim \mathcal{U}(0,1),
\]
with corresponding target velocity
\[
u = a - z .
\]
The action expert $v_\phi$ is trained via the flow-matching objective
\[
\mathcal{L}_{\text{flow}} =
\mathbb{E}\!\left[
\big\| v_\phi(x_\tau,\tau; o, l) - (a - z) \big\|_2^2
\right],
\]
which encourages the model to learn a continuous vector field transporting noise to valid actions conditioned on the observation and task.

At inference time, actions are generated by integrating the learned dynamics
\[
\frac{dx_\tau}{d\tau} = v_\phi(x_\tau,\tau; o, l)
\]
from $\tau=0$ to $\tau=1$ using a fixed-step explicit ODE solver.

\subsection{Training Data} 
The full dataset mixture used for training is summarized in Table~\ref{tab:data_mix}. The reported weights account for each dataset’s relative size and therefore reflect the true fraction of samples drawn from each source within a training batch. We assign a higher proportion to the DROID dataset~\cite{khazatsky2025droidlargescaleinthewildrobot} due to its broad task coverage and rich environmental diversity.

\begin{table}[th]
    \centering
    \caption{Training data mixture. Percentages denote the true fraction of samples from each dataset within each training batch.}
    
    \begin{tabular}{lr}
        \toprule
        \multicolumn{2}{c}{\textbf{Training Dataset Mixture}}\\
        \midrule
        DROID \cite{khazatsky2025droidlargescaleinthewildrobot} & 85.26\% \\
        Fractal \cite{brohan2023rt1roboticstransformerrealworld} & 5.86\% \\
        Bridge \cite{walke2024bridgedatav2datasetrobot} & 3.39\% \\
        BC-Z \cite{pmlr-v164-jang22a} & 0.47\% \\
        Taco Play \cite{mees2023groundinglanguagevisualaffordances} & 0.73\% \\
        Jaco Play \cite{dass2023jacoplay} & 0.12\% \\
        Furniture Bench Dataset \cite{heo2023furniturebenchreproduciblerealworldbenchmark} & 0.31\% \\
        UTAustin Mutex \cite{shah2023mutex} & 0.56\% \\
        Berkeley Fanuc Manipulation \cite{zhu2023fanuc} & 0.19\% \\
        FMB Dataset \cite{luo2024fmbfunctionalmanipulationbenchmark} & 0.09\% \\
        Berkeley Autolab UR5 \cite{BerkeleyUR5Website} & 0.15\% \\
        Austin Buds Dataset \cite{zhu2022bottomupskilldiscoveryunsegmented} & 0.05\% \\
        Austin Sailor Dataset \cite{nasiriany2022sailor} & 0.53\% \\
        Austin Sirius Dataset \cite{liu2023robotlearningjobhumanintheloop} & 0.44\% \\
        Viola \cite{zhu2023violaimitationlearningvisionbased} & 0.12\% \\
        MolmoAct \cite{lee2025molmoactactionreasoningmodels} & 1.73\% \\
        \bottomrule
    \end{tabular}
    \label{tab:data_mix}
\end{table}

During training, we filter idle segments by removing action sequences below a small motion threshold for consecutive steps and discard trajectories lacking task instructions.

\subsection{Compute and Infrastructure Details}
\label{app:compute}

\noindent\textbf{Hero-run compute.}
The final reported checkpoints for all models, including \textsc{LAP-3B}, \textsc{LAP-3B} with VQA co-training, and all replicated baselines, were trained using an identical large-scale pre-training setup. Each hero run was trained for approximately 50 hours on a TPU v6e-64 slice, corresponding to the checkpoints used for all main experiments. Across all reported models, the total hero-run compute amounts to approximately \textbf{50 TPU v6e-64 hours per model}.

\medskip
\noindent\textbf{Total pre-training compute.}
Across the full set of pre-training experiments—including \textsc{LAP-3B} and all replicated baselines, but excluding early prototyping, debugging runs, and downstream fine-tuning experiments—we conducted approximately 200 large-scale training runs. Collectively, these experiments consumed over \textbf{4,000 TPU v6e-64 hours}.

\medskip
\noindent\textbf{Inference compute.}
All real-robot evaluations are executed in real time using a single commodity GPU (an NVIDIA RTX 4090), without any additional model parallelism, specialized hardware, or acceleration techniques. End-to-end real-robot evaluation for the reported experiments required approximately \textbf{24 hours of total human-supervised runtime}.

\section{Evaluation Protocol}
\label{app:evaluation_details}

Each task is evaluated over 20 independent trials with randomized object placement within workspace bounds. A trial is terminated upon task success, time-out, or a safety-triggered abort, such as when the robot collides with the table, reaches joint limits, or exhibits unsafe behavior (e.g., excessively rapid or unstable motions).

For zero-shot experiments, we report binary task success along with 95\% finite-sample valid confidence intervals for binomial success rates, which control the Type-I error (miscoverage) probability \cite{vincent2024generalizablebehaviorcloningpolicy}.

For fine-tuning experiments, we use staged success scores to capture partial task completion:

\noindent$\bullet$ \emph{Hang Tape on Rack:}
\begin{itemize}
\item 0.25: reach the tape
\item 0.5: grasp the tape
\item 0.75: reach the rack while holding the tape
\item 1.0: successfully hang the tape
\end{itemize}

\noindent$\bullet$ \emph{Fold Towel and Place in Basket:}
\begin{itemize}
\item 0.5: correctly folded towel
\item 0.75: placed in basket but unfolded
\item 1.0: full task success
\end{itemize}

\begin{figure*}[t]
    \centering
    \includegraphics[width=\linewidth]{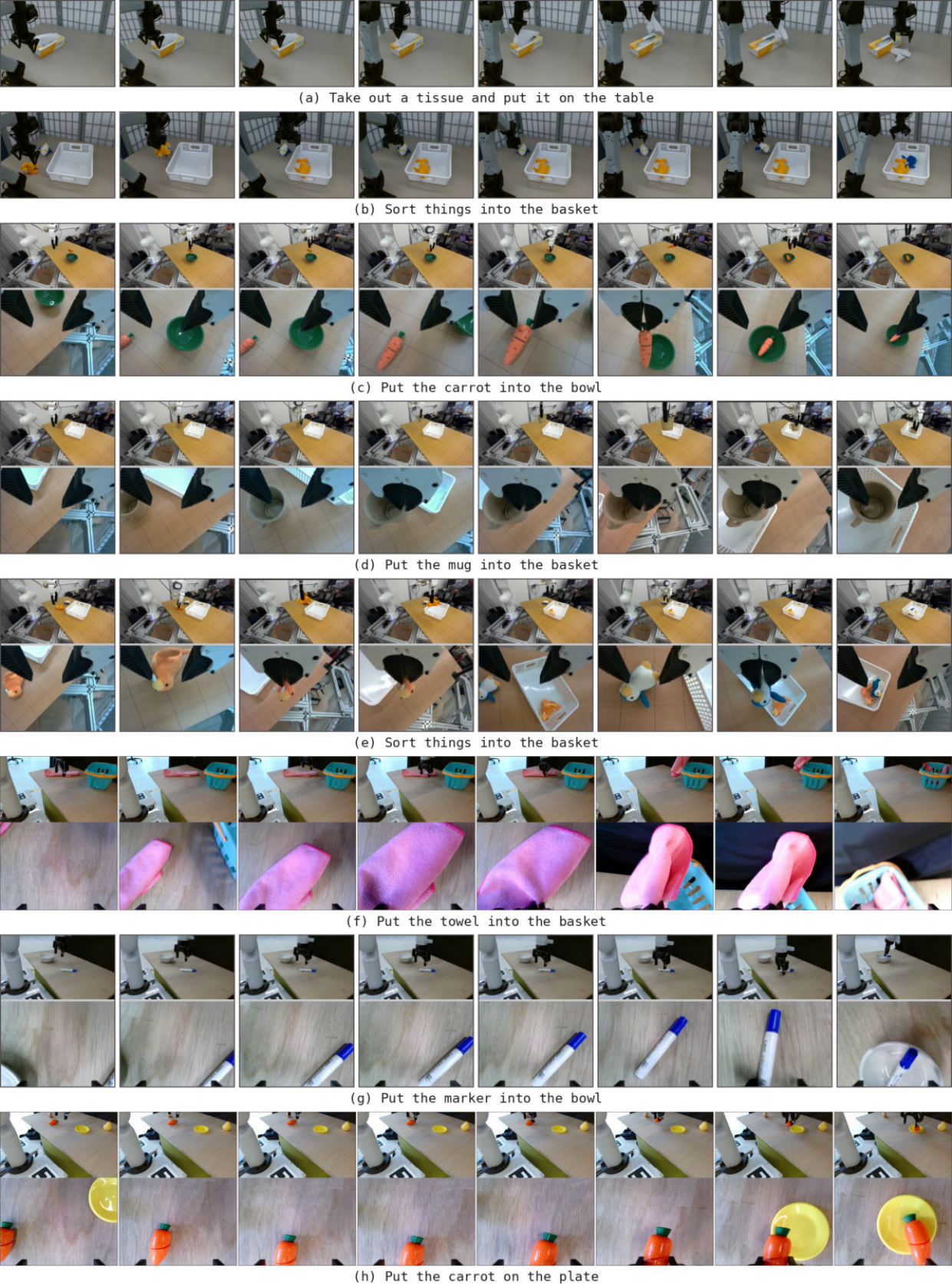}
    \caption{\textbf{Zero-shot evaluation rollout visualizations.} Example observations from the camera views used for policy evaluation across different robot embodiments and tasks, shown from left to right.}
    \label{fig:zeroshot-rollout-visualization}
\end{figure*}

}

\end{document}